\PassOptionsToPackage{table}{xcolor}
\documentclass[11pt]{article}
\usepackage{rotating}
\usepackage{amssymb}

\usepackage[final]{acl}

\usepackage{times}
\usepackage{latexsym}
\usepackage[table]{xcolor}

\newcommand{\dnone}[1]{\cellcolor{gray!8}#1}
\newcommand{\dsmall}[1]{\cellcolor{yellow!25}#1}
\newcommand{\dmedium}[1]{\cellcolor{orange!30}#1}
\newcommand{\dlarge}[1]{\cellcolor{red!35}#1}

\newcommand{\improved}[1]{\cellcolor{blue!12}#1}

\newcommand{\featgroup}[2]{%
  \addlinespace[2pt]
  \multicolumn{2}{@{}l}{%
    \textbf{#1}\enspace\textit{(#2 predictors)}
  }\\[-0.25ex]
}
\usepackage[T1]{fontenc}

\usepackage{comment}

\usepackage{tabularx}
\usepackage{booktabs}
\usepackage{multirow}
\usepackage{array}

\definecolor{rawfamily}{HTML}{DCEAF7}
\definecolor{uposfamily}{HTML}{FCE8D5}
\definecolor{syntaxfamily}{HTML}{E3F1DF}
\definecolor{verbfamily}{HTML}{EDE3F5}

\newcommand{\featurebox}[2]{%
  \begingroup
  \setlength{\fboxsep}{1.2pt}%
  \colorbox{#1}{\strut\texttt{#2}}%
  \endgroup
}

\newcommand{\rawfeat}[1]{\featurebox{rawfamily}{#1}}
\newcommand{\uposfeat}[1]{\featurebox{uposfamily}{#1}}
\newcommand{\synfeat}[1]{\featurebox{syntaxfamily}{#1}}
\newcommand{\verbfeat}[1]{\featurebox{verbfamily}{#1}}
\newcolumntype{Y}{>{\raggedright\arraybackslash}X}

\newcommand{\familykey}[2]{%
  \begingroup
  \setlength{\fboxsep}{1.5pt}%
  \colorbox{#1}{\strut #2}%
  \endgroup
}

\definecolor{rawtextcolor}{RGB}{220,234,247}
\definecolor{uposcolor}{RGB}{252,232,213}
\definecolor{dependencycolor}{RGB}{227,241,223}
\definecolor{verbalcolor}{RGB}{237,227,245}

\newcommand{\rawfeature}[1]{%
  \colorbox{rawtextcolor}{\strut\texttt{#1}}}
\newcommand{\uposfeature}[1]{%
  \colorbox{uposcolor}{\strut\texttt{#1}}}
\newcommand{\depfeature}[1]{%
  \colorbox{dependencycolor}{\strut\texttt{#1}}}
\newcommand{\verbalfeature}[1]{%
  \colorbox{verbalcolor}{\strut\texttt{#1}}}

\definecolor{featureaccent}{HTML}{006D77}
\definecolor{highidrow}{HTML}{EEF8F5}
\definecolor{lowidrow}{HTML}{FCF2EF}
\definecolor{headerrow}{HTML}{E9EEF2}

\newcommand{\featurehl}[1]{%
  \textcolor{featureaccent}{\bfseries #1}%
}

\definecolor{freqfamily}{RGB}{232,222,245}
\newcommand{\freqfeature}[1]{%
  \colorbox{freqfamily}{\strut\texttt{#1}}}

\newcolumntype{C}[1]{>{\centering\arraybackslash}p{#1}}

\newcommand{\syntaxfeat}[1]{\synfeat{#1}}
\newcommand{\verbalfeat}[1]{\verbfeat{#1}}

\definecolor{frequencyfamily}{RGB}{232,222,245}
\newcommand{\frequencyfeat}[1]{%
  \featurebox{frequencyfamily}{#1}}

\usepackage[utf8]{inputenc}
\newcommand{\appendixnumberline}[1]{Appendix\space}
\let\oldappendix\appendix
\makeatletter
\renewcommand{\appendix}{%
  \addtocontents{toc}{\protect\let\protect\numberline\protect\appendixnumberline}%
  \renewcommand{\@seccntformat}[1]{Appendix~\csname the##1\endcsname.\quad }
  \oldappendix
}
\makeatother

\usepackage{microtype}
\usepackage{inconsolata}
\usepackage{graphicx}
\usepackage{amsmath}


\title{Linguistic Features of Local Geometry in Language Models}

\author{
 \textbf{Lucia Domenichelli\textsuperscript{1,2}},
 \textbf{Iuri Macocco\textsuperscript{3}},
 \textbf{Dominique Brunato\textsuperscript{1}},\\
 \textbf{Felice Dell'Orletta\textsuperscript{1}},
 \textbf{Marco Baroni\textsuperscript{3,4}}
\\
 \textsuperscript{1}Istituto di Linguistica Computazionale ``Antonio Zampolli'' (CNR-ILC), ItaliaNLP Lab, Pisa\\
 \textsuperscript{2}University of Pisa\\
 \textsuperscript{3}Universitat Pompeu Fabra (UPF)\\
 \textsuperscript{4}ICREA\\
  \texttt{\{lucia.domenichelli,dominique.brunato,felice.dellorletta\}@ilc.cnr.it}\\
  \texttt{\{iuri.macocco,marco.baroni\}@upf.edu}
}
\begin{document}
\maketitle
\begin{abstract}
Intrinsic dimensionality (ID) describes the effective complexity of language-model representations, but dataset-level estimates obscure variation across individual inputs and make it difficult to examine multiple linguistic properties jointly. We address this limitation by estimating \textit{pointwise ID} relative to a fixed reference set and relating it to interpretable surface, morphosyntactic, and syntactic features. Across Pythia, Qwen, and Apertus, compact linguistic profiles predict pointwise ID at every analysed depth, with the strongest relationship consistently emerging in intermediate layers. Early-layer geometry is more closely associated with surface properties, whereas nominal composition, pronouns, subject dependencies, and verbal structure become more prominent at intermediate and later depths. These patterns persist across ID estimators, neighbourhood sizes, and controls for input length and lexical frequency. Our findings show that variation in local representation geometry hidden by dataset-level summaries is systematically associated with the linguistic properties of individual inputs.\footnote{Code, derived data, and reproduction materials are available in the \href{https://github.com/ldomenichelli/linguistic-features-of-local-geometry-in-language-models}{project's repo}.}
\end{abstract}

\begin{figure}[t!]
    \centering
    \includegraphics[width=\columnwidth]{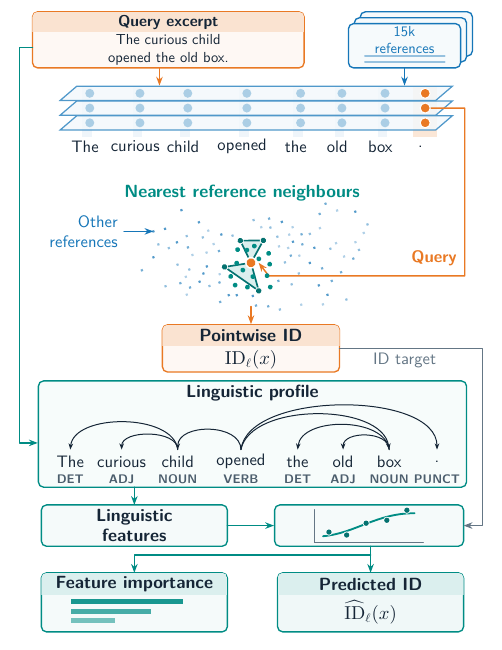}    
\caption{Study pipeline linking an excerpt's linguistic profile to the pointwise ID of its language-model representation. ID is estimated relative to a disjoint reference set and predicted from linguistic features.}
\label{fig:pipeline}
\end{figure}

\section{Introduction}\label{sec:introduction}

Despite their remarkable success across a wide range of tasks, transformer-based language models remain difficult to interpret. While benchmark evaluations reveal what models can accomplish, they provide only indirect evidence on how linguistic information is represented internally. We examine this internal organization by asking how the geometry of a text's representation relates to its linguistic properties. Previous work provides a basis for this question: linguistic features have been found to occupy hierarchically organized low-dimensional subspaces \citep{hernandez-andreas-2021-low}, and syntactic structure has been shown to be encoded in the geometry of representation spaces \citep{diego-simon-etal-2024-polar}.

Intrinsic dimensionality (ID) describes one aspect of this geometry by estimating the number of locally independent directions along which representations vary. It provides a compact characterization of local representational complexity: a lower estimate indicates variation along fewer independent directions, and a higher estimate indicates variation along more. Previous studies have shown that ID changes systematically across transformer layers and varies with linguistic constructions, model architectures, and text characteristics.

Most studies, however, characterize ID at the level of an entire dataset or a predefined subset of representations \cite{cheng-etal-2023-bridging,domenichelli-etal-2026-linguistic}. These aggregate estimates reveal broad geometric trends, but obscure differences between individual inputs. Studying a particular linguistic property in this way generally requires dividing the data into matched subsets and estimating ID separately for each condition. This is manageable for a few controlled contrasts, but becomes cumbersome when properties are continuous or when several must be considered together. The aggregate approach therefore leaves open which properties of a particular input are associated with the geometric complexity of its representation.

\textit{Pointwise} ID addresses this gap by assigning each representation an estimate of the dimensionality of its local neighbourhood. This allows us to model geometric variation directly as a function of individual inputs' linguistic properties. We ask three related questions: (i) Can the local geometry of individual representations be predicted from their linguistic properties? (ii) Which linguistic properties are most consistently associated with variation in pointwise ID? (iii) How do these associations evolve across model depth and during pretraining?

We study these questions in three decoder-only models: Pythia, Qwen, and Apertus. Each text excerpt is represented by the hidden state of its final non-padding token. At selected early, intermediate, and late layers, we estimate its pointwise ID using neighbouring representations from a fixed set of reference excerpts, disjoint from the evaluation excerpts. We predict these estimates from compact linguistic profiles spanning surface, morphosyntactic, and syntactic properties, allowing several properties to be considered jointly for each excerpt. We also examine available Pythia and Apertus checkpoints to trace changes during pretraining.

Across all three models, these linguistic profiles predict pointwise ID at every examined depth, with the strongest predictive performance in intermediate layers. We identify the features associated with this variation, track how their roles change across depth and pretraining, and assess robustness across estimators and experimental settings. These results show that linguistic differences between individual inputs are systematically associated with local geometric variation in the models and neighbourhood settings examined.

\section{Related Work}\label{sec:related}
Language model representations live in spaces with hundreds or thousands of coordinates, but the data need not vary independently along all of them. Instead, they may concentrate near a structure with far fewer effective degrees of freedom. Intrinsic dimensionality (ID) seeks to recover the dimension of this underlying structure \citep{fukunaga1971algorithm}. 
Linear approaches assume that the data can be approximated by a single linear or affine subspace. For example, PCA-based estimates count the principal components required to explain a specified proportion of the variance \citep{pearson1901lines}. Nonlinear approaches instead allow the data to lie near a curved manifold and infer its dimension from distances or geometric relations among neighbouring observations \citep{levina2004maximum,facco2017estimating}.
Most linguistic studies of ID assign a single estimate to an entire dataset, model layer, or experimentally defined condition. \citet{cheng-etal-2023-bridging} connected geometric ID to language-model coding length and ease of adaptation, while showing that properties such as vocabulary size and syntactic structure modulate dimensionality. \citet{cheng2025emergence} subsequently identified an intermediate high-dimensional phase associated with abstract linguistic representations, downstream transfer, and cross-model alignment.
Controlled comparisons have further shown that coordination and subordination, branching and syntactic embedding, and attachment ambiguity produce distinct layer-wise ID profiles \citep{baroni-etal-2026-tracing}. Similarly, \citet{domenichelli-etal-2026-linguistic} partitioned token representations according to surface and syntactic properties and found systematic differences in isotropy and in both linear and nonlinear ID. Together, these studies demonstrate that aggregate dimensionality profiles reflect the linguistic organization of the represented data. Because they summarize predefined groups, however, they do not directly characterize how dimensionality varies from one observation to another as a function of multiple continuous and correlated linguistic properties.

A related line of work has moved toward finer-grained units of analysis.
\citet{tulchinskii-etal-2023-intrinsic} represented each text as a cloud of contextualized token embeddings and used the persistent-homology dimension of that cloud to distinguish human- from machine-generated text. \citet{pedashenko-etal-2026-unveiling} subsequently related such text-level estimates to lexical, stylistic, and genre properties. Although these approaches produce one estimate for each text, that estimate describes the geometry of the token cloud \emph{within} the text. It is therefore different from estimating the local dimension around the text's representation within a shared space of excerpts. 
A genuinely pointwise formulation has been used for a different purpose. \citet{NEURIPS2025_61c2c633} estimated local dimensions for contextual token representations and then analysed their aggregate evolution during training and fine-tuning. Their results connected changes in mean local dimension to grokking, overfitting, convergence, and generalization, rather than to the linguistic properties of individual inputs.

Our work connects these strands by treating every excerpt as one point in a shared representation space and retaining its pointwise ID as the quantity of interest. For each excerpt, we estimate ID from its neighbourhood in a fixed, disjoint reference set, rather than treating its tokens as an independent point cloud or assigning a single dimension to a predefined linguistic condition. We then ask which
continuous and correlated surface, morphological, and syntactic properties predict this variation. Tracking these relationships across model depth and pretraining allows us to examine how the linguistic correlates of local representational geometry emerge and change.
\section{Methods}\label{sec:methods}
Our framework combines an input-specific geometric descriptor of language model representations with a structured linguistic characterization of the corresponding text. Given an LLM and a collection of text excerpts, we first estimate the pointwise intrinsic dimensionality of each excerpt from the geometry of its local neighbourhood, yielding an input-specific pointwise ID score. We then investigate whether this score can be predicted from the linguistic profile of the corresponding excerpt. This framework is applied across models to assess whether the observed relationships generalize across model families, across layers to identify where they are strongest, and across intermediate pretraining checkpoints to trace how they emerge and evolve during pretraining.

\subsection{Geometric analysis}\label{sec:implementation_details}
\paragraph{Models and representations}

We analyse Pythia-6.9B-deduped \citep{pmlr-v202-biderman23a}, Qwen2.5-7B \citep{qwen2025technical}, and Apertus-8B \citep{hernandez-cano-etal-2026-apertus}. For each model and excerpt, we extract the hidden state of the final non-padding token at one early, one intermediate, and one late layer.

\paragraph{Reference and evaluation data}\label{par:dataset}
Let \(h_{\ell}(s)\in\mathbb{R}^{d_\ell}\) denote the representation of an excerpt \(s\) at layer \(\ell\). At each layer, we construct a fixed reference set
\[
\mathcal{R}_{\ell}
=
\{h_{\ell}(r_1),\ldots,h_{\ell}(r_{n_{\mathrm{ref}}})\},
\]
where the reference excerpts are disjoint from the evaluation ones. Given an evaluation excerpt \(x\), we compute its Euclidean distance to every representation in \(\mathcal{R}_{\ell}\) and retrieve the \(k\) nearest reference neighbours. We denote the resulting ordered distances by
\[
T_1(x) \leq T_2(x) \leq \cdots \leq T_k(x),
\]
where \(T_j(x)\) is the distance from \(h_{\ell}(x)\) to its \(j\)-th nearest reference neighbour. We write
\(\mathcal{N}_{\ell,k}(x)\) for the corresponding cloud of neighbour representations. The target-to-neighbour distances and the internal geometry of this cloud provide the local information used by the ID estimators. We selected \(n_{\mathrm{ref}}\) and \(k\) through preliminary sweeps over reference-set and neighbourhood sizes. Small neighbourhoods produced unstable estimates, whereas enlarging the reference set beyond a certain size changed the pointwise scores only marginally (App.~\ref{app:reference_size}, \ref{app:stability_reference}, and~\ref{app:k}). These results motivated the fixed operating point used throughout the main analysis. 

We chose \(k=1000\) primarily to obtain stable ID estimates. Sensitivity analyses for Pythia and Qwen indicate broadly consistent depth-dependent patterns across the neighbourhood sizes examined (App.~\ref{app:extra_analysis}), while predictive performance varies with \(k\). These findings support the robustness of the main patterns across the tested settings. We also recognise that stability does not necessarily imply geometric locality, and interpret our estimates in relation to the chosen neighbourhood scale.

The reference set ensures that we estimate the neighbourhood around each evaluation target based on a stable set of reference points randomly sampled from the set of possible excerpts. Note that previous works on ID estimation in LLMs use the same items as both reference points and queries, making the estimation overly reliant on the chosen evaluation set.

We sample both the reference and the evaluation set from the training split of The Pile \citep{gao2020pile}, the original Pythia pretraining corpus. Both sets undergo the same preprocessing: excerpts contain at most 100 tokens (according to the Pythia tokenizer) and are truncated at a period boundary.
As an additional robustness check, Appendix~\ref{app:extra_results} shows that the marginal distribution of pointwise ID is similar for evaluation excerpts sampled from the Pile's training and test splits.

\paragraph{Pointwise ID estimation}
Pointwise intrinsic dimensionality assigns each excerpt an estimate of
the effective dimensionality of its local neighbourhood. Following the
Local Intrinsic Dimensionality framework \citep{houle2017local}, we derive this estimate from the
local distribution of distances around the excerpt, rather than assigning a
single dimension to the entire representation space. 

We use Expected Simplex Skewness (ESS) as the primary estimator \citep{johnsson2014low, bac2021scikit}. 
For each query item \(x\), we apply ESS separately to the local cloud \(\mathcal{N}_{\ell,k}(x)\) formed by its \(k\) nearest reference
representations at layer \(\ell\):
\[
\operatorname{ID}^{(\mathrm{ESS})}_{\ell}
    (x;k,n_{\mathrm{ref}})
=
\widehat{m}_{\mathrm{ESS}}
    \!\left(\mathcal{N}_{\ell,k}(x)\right).
\]
ESS estimates effective dimensionality from the angular and volumetric structure of simplices formed within this local cloud. Intuitively, a neighbourhood concentrated along relatively few independent directions yields a lower estimate, whereas a neighbourhood extending across more independent directions yields a higher estimate.

For robustness, we repeat the analyses with two complementary estimators: the Full Correlation Integral estimator (FCI) \citep{erba2019intrinsic} and a maximum-likelihood estimator (MLE) \citep{levina2004maximum}. These estimators rely on different aspects of neighbourhood geometry and are described, together with their implementation details and calibration analyses, in App.~\ref{app:estimators}, ~\ref{app:k} and~\ref{app:extra_analysis}.

\subsection{Linguistic Prediction Framework} \label{par:linguistic_features}
\paragraph{Linguistic Features}

We represent each excerpt using 43 descriptors drawn from Profiling-UD \citep{brunato-etal-2020-profiling}, a multilingual tool for linguistic profiling based on the Universal Dependencies (UD) annotation framework. These descriptors account for a broad range of linguistic phenomena spanning multiple levels of linguistic analysis, including surface, lexical, and (morpho-)syntactic properties. We select a compact subset of the full inventory that is appropriate for relatively short, individual inputs, preserving broad linguistic coverage while emphasizing features that can be clearly interpreted at the excerpt level.

We group these descriptors into four broad feature families\footnote{The full list of Profiling-UD features is provided in Table~\ref{tab:retained_profiling_ud_features} in Appendix~\ref{app:features}.}. \emph{Raw text} features capture surface-level properties of the input, including the numbers of sentences and tokens, average sentence length, and average token length in characters. \emph{Morphosyntactic} features characterize the distribution of part-of-speech (POS) categories defined in the UD annotation scheme, as well as lexical density. \emph{Dependency and tree structure} features summarize the distribution of UD dependency labels, the average length of dependency links and the maximum depth of the parsed tree, as well as linear-order properties (e.g. dependent subjects preceding or following their verbal head). Finally, \emph{Verbal and clausal structure} features characterize clause organization and verbal predicates and they include the distribution of subordinate clauses and nominal structures headed by a preposition, the distribution of verbal roots and inflectional UD features (such as tense, mood, and number) for lexical verbs and auxiliaries, and verb arity (calculated as the number of instantiated dependency links sharing the same verbal head). This classification helps relate patterns in feature importance to broader linguistic properties rather than isolated descriptors.

To distinguish these structural features from lexical-frequency effects on ID, we add three controls computed using each model's native tokenizer. Token frequencies are estimated on the same reference corpus, The Pile \citep{gao2020pile}, and converted into logarithmic frequency classes. We denote the class of token \(t\) under model \(M\)'s tokenizer by \(C_M(t)\).
For each excerpt, we retain the median, maximum, and minimum classes:  \texttt{\{median,max,min\}\_frequency\_class}.

\paragraph{Prediction tasks}\label{sec:eval}
For each model and layer, we paired each excerpt's 46-dimensional feature vector, comprising linguistic profiling descriptors and lexical frequency features, with its corresponding ID score. For continuous prediction, we used random forest regression and report mean five-fold cross-validated $R^2$ (hyperparameters and cross-validation details are provided in App.~\ref{app:hyp}). As a complementary analysis, we used random forest classifiers to distinguish excerpts from the lowest and highest ID quintiles (App.~\ref{app:classification}).

\section{Results}\label{sec:results}
\subsection{Layer-wise ID profiles}\label{sec:profiles}
\begin{figure}[t!]
    \centering
    \includegraphics[width=\linewidth]{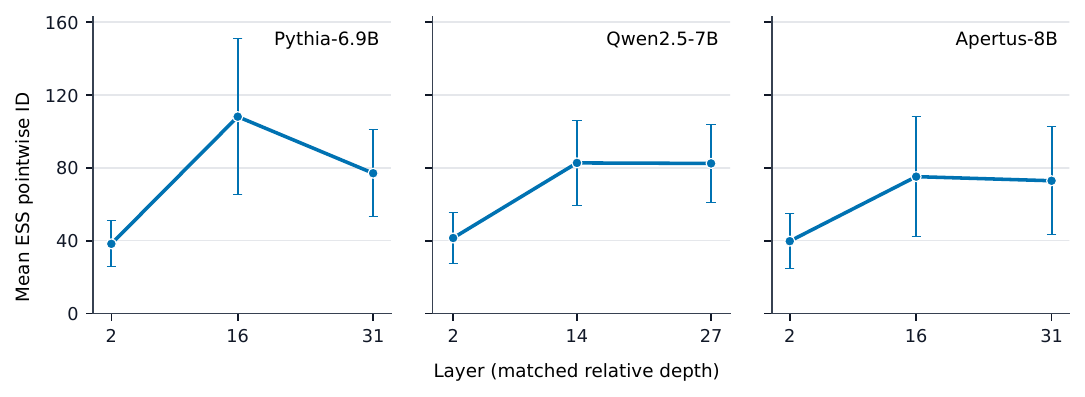}
    \caption{Mean ESS-estimated pointwise ID (\(\pm1\sigma\), where \(\sigma\) is the standard deviation across excerpts) across approximately matched relative depths in Pythia, Qwen, and Apertus.}
    \label{fig:ess_profile}
\end{figure}

Unless otherwise stated, the main experiments use $n_{\mathrm{ref}}=15{,}000$ and $k=1000$. 
We first examine aggregate layer-wise geometry. Averaged across evaluation excerpts, pointwise ID rises from early to intermediate depth in all three models and then decreases or stabilizes toward the final layers (Figure~\ref{fig:ess_profile}), consistent with reports of a central high-dimensional phase in language transformers \citep{cheng2025emergence}. However, broad within-layer dispersion shows that these means conceal substantial variation, motivating our central question: whether input-level geometry systematically relates to linguistic form.

As a control analysis, we also examine the relationship between pointwise ID and perplexity, asking whether local dimensionality primarily reflects model surprisal (Table \ref{tab:perplexity_ess_correlations}). For Pythia and Qwen, perplexity is moderately and positively correlated with pointwise ID, with the strongest association at intermediate depth. Apertus shows a weaker and less consistent relationship across layers. Here too, the strongest correlation occurs at the intermediate layer. It is particularly intriguing that the intermediate layer, sometimes argued to correspond to the phase of deepest linguistic analysis, is more strongly associated with perplexity than the late layer, which is closer to the output prediction step. Overall, though, the results indicate that model surprisal and local dimensionality capture partially overlapping but non-equivalent input properties.

\subsection{Predicting pointwise ID from linguistic profiles}
\label{par:linguistic_profiling}

Across all three model families and at every analysed layer, the full linguistic profiles predicted ID with positive cross-validated \(R^2\) (Table~\ref{tab:main-results}). The corresponding predictive results for MLE and FCI are reported in Appendix~\ref{app:id_estimator_robustness}. For the random forest regressor, performance was highest at intermediate depth in every model, coinciding with the highest mean ID among the analysed layers in Figure~\ref{fig:ess_profile}. Linguistic properties are therefore most predictive of local geometric variation at this stage of representation processing. 
The same qualitative pattern emerged in the complementary binary classification task, whose performance likewise peaked at intermediate depth (App.~\ref{app:classification}), reinforcing the conclusion that linguistic information is most predictive of local representation geometry at this stage of processing.

\begin{table}[t!]
\centering
\footnotesize
\setlength{\tabcolsep}{2.5pt}
\renewcommand{\arraystretch}{1.10}
\renewcommand{\tabularxcolumn}[1]{m{#1}}

\begin{tabularx}{\columnwidth}{
    @{}l *{3}{>{\centering\arraybackslash}X}@{}
}
\toprule
\textbf{Model}
& \textbf{Early}
& \textbf{Intermediate}
& \textbf{Late} \\
\midrule

Pythia-6.9B
& 2: .088
& \textbf{16: .373}
& 31: .299 \\

Qwen2.5-7B
& 2: .095
& \textbf{14: .490}
& 27: .348 \\

Apertus-8B
& 2: -.097
& \textbf{16: .159}
& 31: .124 \\
\bottomrule
\end{tabularx}
\caption{Spearman correlations between excerpt perplexity and ESS-estimated pointwise ID. Each cell reports ``layer: \(\rho\)''; boldface marks each model's largest absolute correlation. All correlations satisfy \(p<.050\).}
\label{tab:perplexity_ess_correlations}
\end{table}

\begin{table*}[h!]
\centering
\small
\setlength{\tabcolsep}{4pt}
\renewcommand{\arraystretch}{1.12}
\renewcommand{\tabularxcolumn}[1]{m{#1}}

\begin{tabularx}{\textwidth}{
  @{}l c *{6}{>{\centering\arraybackslash}X}@{}
}
\toprule
\multirow{2}{*}{\textbf{Model}}
& \multirow{2}{*}{\textbf{Layer}}
& \multicolumn{3}{c}{\textbf{Linguistic features}}
& \multicolumn{3}{c}{\textbf{Baselines}} \\
\cmidrule(lr){3-5}
\cmidrule(l){6-8}
& & \textbf{Full}
& \textbf{\shortstack[c]{No\\frequency}}
& \textbf{\shortstack[c]{No raw\\features}}
& \textbf{\shortstack[c]{TF-IDF\\frequent}}
& \textbf{\shortstack[c]{TF-IDF\\1-5\%}}
& \textbf{\shortstack[c]{Log\\perplexity}} \\
\midrule

\multirow{3}{*}{Pythia-6.9B}
& 2  & \textbf{.256} & .254 & .215 & .154 & .068 & -.292 \\
& 16 & \textbf{.505} & .504 & .476 & .274 & .221 & -.077 \\
& 31 & \textbf{.263} & .262 & .257 & .135 & .104 & -.133 \\

\addlinespace

\multirow{3}{*}{Qwen2.5-7B}
& 2  & \textbf{.246} & .244 & .199 & .098 & -.017 & -.266 \\
& 14 & \textbf{.547} & \textbf{.547} & .527 & .323 & .269 & .006 \\
& 27 & \textbf{.432} & .430 & .419 & .239 & .180 & -.105 \\

\addlinespace

\multirow{3}{*}{Apertus-8B}
& 2  & .270 & \textbf{.271} & .206 & .130 & -.031 & -.063 \\
& 16 & \textbf{.439} & \textbf{.439} & .416 & .258 & .199 & -.075 \\
& 31 & \textbf{.311} & .307 & .291 & .193 & .121 & -.114 \\

\bottomrule
\end{tabularx}

\caption{ESS-estimated pointwise ID regression with linguistic features, ablations, and baselines. Values are mean five-fold cross-validated \(R^2\) scores for random forest regression using the same excerpts and folds within each model-layer setting. \textit{Full} uses all 46 predictors; \textit{No frequency} removes the three frequency controls, while \textit{No raw features} removes the four surface descriptors. Each TF-IDF baseline uses 46 word features; the log-perplexity baseline uses one feature. Boldface marks the highest displayed score in each row, including ties.}
\label{tab:main-results}
\end{table*}

Table~\ref{tab:main-results} also reports two sets of controls. 
First, we ablated lexical-frequency and raw-text features from the linguistic profile. Removing the three frequency controls had a negligible effect on performance, whereas removing the four raw features (\texttt{n\_sentences}, \texttt{n\_tokens}, \texttt{tokens\_per\_sent}, and \texttt{char\_per\_tok}) produced larger reductions in \(R^2\), particularly at early depth. In all cases, however, the intermediate-layer advantage remained.

Second, we compared the linguistic profiling-based features with lexical and model-surprisal baselines. 
The full linguistic profile outperformed both dimension-matched TF-IDF baselines based on frequent unigrams and on unigrams occurring in 1-5\% of the training excerpts, as well as the log-perplexity baseline in all nine model-layer settings. In particular, a random forest using log perplexity as its only predictor yielded near-zero or negative cross-validated \(R^2\), suggesting that the linguistic profile contains predictive information beyond that represented by these lexical and model-surprisal baselines. 

These controls do not isolate linguistic structure from every possible lexical- or frequency-based confound, but they show that the observed predictability is not reducible to token frequency or surface features alone. Importantly, they also show \textit{where} in the model such factors are more or less prominent.
The full regression and classification comparisons across ID estimators are reported in Tables~\ref{tab:lexical_regression} and~\ref{tab:lexical_classification} in App.~\ref{app:lexical}.

\subsection{Predictive features across models and depth}
Having established the overall relationship between linguistic features and pointwise ID in the previous section, we next examine which linguistic phenomena recur across model families, and how their relevance changes with depth. We use held-out permutation importance to identify the predictors on which the fitted random forests rely most. 

The rankings in Table~\ref{tab:top-features} identify a compact set of features that recur across models. Nominal composition is particularly prominent: proper- and common-noun proportions (\texttt{upos\_PROPN} and \texttt{upos\_NOUN}) repeatedly appear among the leading predictors. Average token length (\texttt{char\_per\_tok}), pronoun proportion (\texttt{upos\_PRON}), nominal-subject dependencies (\texttt{dep\_nsubj}), verbal-root prevalence (\texttt{verbal\_root}), token count (\texttt{n\_tokens}), and punctuation proportion (\texttt{upos\_PUNCT}) also recur. The predictive profile therefore spans token form, grammatical composition, and predicate-argument organization rather than being dominated by a single feature family.

Other feature families show less stable evidence under the same ranking criterion. Measures of dependency-tree length and depth and most fine-grained verbal and auxiliary morphology do not recur among the leading predictors. Consistent with the ablation results in Section~\ref{par:linguistic_profiling}, neither do the tokenizer-specific frequency summaries.

The importance profiles show a redistribution of predictive information. At early layers, token count is the leading predictor for all three models, while punctuation and average characters per token also recur. At intermediate depth, common- and proper-noun proportions and average characters per token appear among the five leading predictors for every model. Common-noun proportion is the leading predictor for Pythia and Apertus, whereas Qwen places greatest weight on average characters per token, followed by pronouns and nominal-subject dependencies. At late depth, no
single feature leads across architectures. Pythia is led by common-noun proportion, Qwen by verbal-root prevalence, and Apertus by proper-noun proportion. Nevertheless, proper nouns, nominal-subject dependencies, and verbal roots occur in all three late-layer profiles.

Depth therefore appears to reweight, rather than replace, different sources
of information. Early geometry mainly reflects input form, while intermediate and late geometry increasingly reflects how texts organize entities and events. Dense, nominal technical passages may form more regular local neighbourhoods, whereas narrative passages with more participants and predicates may require a broader range of contextual distinctions. Surface and grammatical information nevertheless remain relevant throughout the model.

\begin{table*}[h!]
\centering
\footnotesize
\setlength{\fboxsep}{0.8pt}
\renewcommand{\arraystretch}{1.08}

\begin{tabularx}{\textwidth}{
    @{}
    c@{\hspace{7pt}}
    >{\raggedright\arraybackslash}X
    @{}
}
\toprule
\textbf{Layer}
& \textbf{Five largest held-out RF permutation importances
\({(\Delta R^2)}\)} \\
\midrule
\rowcolor{black!5}

\multicolumn{2}{@{}l}{\textbf{Pythia-6.9B}} \\

2
& \rawfeat{n\_tokens} (.159),
  \uposfeat{upos\_PROPN} (.041),
  \uposfeat{upos\_PUNCT} (.033),
  \synfeat{dep\_nsubj} (.020),
  \rawfeat{char\_per\_tok} (.018) \\

16
& \uposfeat{upos\_NOUN} (.149),
  \rawfeat{char\_per\_tok} (.091),
  \uposfeat{upos\_PRON} (.074),
  \uposfeat{upos\_PROPN} (.067),
  \verbfeat{verbal\_root} (.044) \\

31
& \uposfeat{upos\_NOUN} (.075),
  \uposfeat{upos\_PROPN} (.061),
  \verbfeat{verbal\_root} (.040),
  \uposfeat{upos\_PRON} (.032),
  \synfeat{dep\_nsubj} (.020) \\

\addlinespace[3pt]
\rowcolor{black!5}

\multicolumn{2}{@{}l}{\textbf{Qwen2.5-7B}} \\

2
& \rawfeat{n\_tokens} (.163),
  \verbfeat{verbal\_root} (.042),
  \uposfeat{upos\_PUNCT} (.033),
  \rawfeat{char\_per\_tok} (.019),
  \uposfeat{upos\_NUM} (.015) \\

14
& \rawfeat{char\_per\_tok} (.136),
  \uposfeat{upos\_PRON} (.107),
  \synfeat{dep\_nsubj} (.053),
  \uposfeat{upos\_PROPN} (.047),
  \uposfeat{upos\_NOUN} (.030) \\

27
& \verbfeat{verbal\_root} (.138),
  \uposfeat{upos\_NOUN} (.125),
  \uposfeat{upos\_PROPN} (.123),
  \synfeat{dep\_nsubj} (.029),
  \rawfeat{char\_per\_tok} (.027) \\

\addlinespace[3pt]
\rowcolor{black!5}

\multicolumn{2}{@{}l}{\textbf{Apertus-8B}} \\

2
& \rawfeat{n\_tokens} (.199),
  \verbfeat{verbal\_root} (.032),
  \rawfeat{n\_sentences} (.019),
  \uposfeat{upos\_PUNCT} (.012),
  \rawfeat{char\_per\_tok} (.010) \\

16
& \uposfeat{upos\_NOUN} (.137),
  \synfeat{dep\_nsubj} (.054),
  \rawfeat{char\_per\_tok} (.039),
  \uposfeat{upos\_PROPN} (.037),
  \verbfeat{verbal\_root} (.032) \\

31
& \uposfeat{upos\_PROPN} (.148),
  \synfeat{dep\_nsubj} (.055),
  \rawfeat{n\_tokens} (.040),
  \verbfeat{verbal\_root} (.034),
  \uposfeat{upos\_PRON} (.026) \\

\bottomrule
\end{tabularx}

\smallskip
\caption{Most predictive features for ESS-estimated pointwise ID regression. Parentheses report the mean decrease in held-out \(R^2\) after permutation, averaged over eight shuffles and five cross-validation folds. Background colours indicate the four linguistic feature families: \familykey{rawfamily}{Raw text}, \familykey{uposfamily}{Morphosyntax}, \familykey{syntaxfamily}{Dependency and tree structure}, and \familykey{verbfamily}{Verbal and clausal structure}.}
\label{tab:top-features}
\end{table*} 

\begin{table}[t]
\centering
\footnotesize
\setlength{\tabcolsep}{2.5pt}
\setlength{\fboxsep}{1pt}
\renewcommand{\arraystretch}{1.10}
\renewcommand{\tabularxcolumn}[1]{m{#1}}

\begin{tabularx}{\columnwidth}{@{}l *{3}{>{\centering\arraybackslash}X}@{}}
\toprule
\textbf{Feature} & \textbf{Early} & \textbf{Middle} & \textbf{Late} \\
\midrule
\rawfeature{n\_tokens}
& \(\uparrow\,3\) & - & \(\uparrow\,1\) \\
\rawfeature{char\_per\_tok}
& \(\uparrow\,2\;\downarrow\,1\) & \(\downarrow\,3\) & \(\downarrow\,1\) \\
\uposfeature{upos\_NOUN}
& - & \(\downarrow\,3\) & \(\downarrow\,2\) \\
\uposfeature{upos\_PRON}
& - & \(\uparrow\,2\) & \(\uparrow\,2\) \\
\uposfeature{upos\_PROPN}
& \(\uparrow\,1\) & \(\uparrow\,3\) & \(\uparrow\,3\) \\
\depfeature{dep\_nsubj}
& \(\uparrow\,1\) & \(\uparrow\,2\) & \(\uparrow\,3\) \\
\verbalfeature{verbal\_root}
& \(\uparrow\,2\) & \(\uparrow\,2\) & \(\uparrow\,3\) \\
\bottomrule
\end{tabularx}

\caption{Stable ALE directions for recurring predictors of ESS-estimated pointwise ID. Arrows indicate positive (\(\uparrow\)) or negative (\(\downarrow\)) ALE contrasts between the 10th and 90th percentiles of a feature. Counts show how many of the three models share each stable direction. Dashes indicate features not selected at that depth. Background colours indicate the four linguistic feature families: \familykey{rawfamily}{Raw text}, \familykey{uposfamily}{Morphosyntax}, \familykey{syntaxfamily}{Dependency and tree structure}, and \familykey{verbfamily}{Verbal and clausal structure}.}
\label{tab:feature_direction}
\end{table}

\begin{table*}[!tp]
\centering
{\scriptsize
\setlength{\tabcolsep}{4pt}
\renewcommand{\arraystretch}{1.02}

\begin{tabularx}{\textwidth}{
    @{}
    l
    c
    r
    >{\raggedright\arraybackslash}X
    @{}
}
\toprule
\textbf{Checkpoint}
& \textbf{Layer}
& \textbf{ESS \(R^2\)}
& \textbf{Five largest permutation importances \((\Delta R^2)\)} \\
\midrule

\multirow{3}{*}{210B}
& 2 & \textbf{.417}
& \rawfeat{n\_tokens} (.255);
  \rawfeat{n\_sentences} (.113);
  \rawfeat{char\_per\_tok} (.026);
  \syntaxfeat{dep\_nsubj} (.023);
  \uposfeat{upos\_PUNCT} (.018) \\

& 16 & \textbf{.446}
& \uposfeat{upos\_NOUN} (.129);
  \uposfeat{upos\_PROPN} (.090);
  \rawfeat{char\_per\_tok} (.050);
  \verbalfeat{verbal\_root} (.047);
  \uposfeat{upos\_PRON} (.046) \\

& 31 & \textbf{.362}
& \uposfeat{upos\_PROPN} (.174);
  \syntaxfeat{dep\_nsubj} (.059);
  \verbalfeat{verbal\_root} (.058);
  \uposfeat{upos\_PRON} (.040);
  \uposfeat{upos\_NOUN} (.022) \\

\addlinespace[2pt]

\multirow{3}{*}{420B}
& 2 & .252
& \rawfeat{n\_tokens} (.188);
  \rawfeat{n\_sentences} (.043);
  \uposfeat{upos\_PUNCT} (.033);
  \rawfeat{lexical\_density} (.008);
  \rawfeat{char\_per\_tok} (.008) \\

& 16 & .431
& \uposfeat{upos\_NOUN} (.112);
  \uposfeat{upos\_PROPN} (.071);
  \verbalfeat{verbal\_root} (.054);
  \rawfeat{char\_per\_tok} (.041);
  \syntaxfeat{dep\_nsubj} (.039) \\

& 31 & .346
& \uposfeat{upos\_PROPN} (.183);
  \verbalfeat{verbal\_root} (.066);
  \syntaxfeat{dep\_nsubj} (.050);
  \rawfeat{n\_tokens} (.027);
  \rawfeat{char\_per\_tok} (.023) \\

\addlinespace[2pt]

\multirow{3}{*}{840B}
& 2 & .274
& \rawfeat{n\_tokens} (.211);
  \uposfeat{upos\_PUNCT} (.043);
  \rawfeat{n\_sentences} (.036);
  \rawfeat{char\_per\_tok} (.014);
  \uposfeat{upos\_PRON} (.010) \\

& 16 & .408
& \uposfeat{upos\_NOUN} (.119);
  \uposfeat{upos\_PROPN} (.066);
  \verbalfeat{verbal\_root} (.043);
  \syntaxfeat{dep\_nsubj} (.042);
  \rawfeat{char\_per\_tok} (.038) \\

& 31 & .315
& \uposfeat{upos\_PROPN} (.171);
  \verbalfeat{verbal\_root} (.046);
  \syntaxfeat{dep\_nsubj} (.034);
  \rawfeat{n\_tokens} (.031);
  \uposfeat{upos\_PRON} (.029) \\

\addlinespace[2pt]

\multirow{3}{*}{1.68T}
& 2 & .359
& \rawfeat{n\_sentences} (.141);
  \rawfeat{n\_tokens} (.125);
  \syntaxfeat{dep\_nsubj} (.029);
  \rawfeat{char\_per\_tok} (.025);
  \uposfeat{upos\_PUNCT} (.017) \\

& 16 & .428
& \uposfeat{upos\_NOUN} (.118);
  \uposfeat{upos\_PROPN} (.071);
  \verbalfeat{verbal\_root} (.057);
  \rawfeat{char\_per\_tok} (.046);
  \uposfeat{upos\_PRON} (.045) \\

& 31 & .349
& \uposfeat{upos\_PROPN} (.169);
  \verbalfeat{verbal\_root} (.062);
  \syntaxfeat{dep\_nsubj} (.055);
  \uposfeat{upos\_PRON} (.032);
  \rawfeat{char\_per\_tok} (.028) \\

\addlinespace[2pt]

\multirow{3}{*}{3.57T}
& 2 & .357
& \rawfeat{n\_tokens} (.220);
  \rawfeat{n\_sentences} (.052);
  \syntaxfeat{dep\_nsubj} (.023);
  \uposfeat{upos\_PUNCT} (.020);
  \rawfeat{char\_per\_tok} (.018) \\

& 16 & .441
& \uposfeat{upos\_NOUN} (.145);
  \uposfeat{upos\_PROPN} (.065);
  \verbalfeat{verbal\_root} (.050);
  \uposfeat{upos\_PRON} (.039);
  \syntaxfeat{dep\_nsubj} (.035) \\

& 31 & .354
& \uposfeat{upos\_PROPN} (.176);
  \verbalfeat{verbal\_root} (.058);
  \syntaxfeat{dep\_nsubj} (.044);
  \uposfeat{upos\_NOUN} (.029);
  \rawfeat{n\_tokens} (.024) \\

\addlinespace[2pt]

\multirow{3}{*}{7.232T}
& 2 & .285
& \rawfeat{n\_tokens} (.191);
  \uposfeat{upos\_PUNCT} (.025);
  \rawfeat{n\_sentences} (.024);
  \syntaxfeat{dep\_nsubj} (.015);
  \uposfeat{upos\_NOUN} (.010) \\

& 16 & .436
& \uposfeat{upos\_NOUN} (.139);
  \verbalfeat{verbal\_root} (.075);
  \uposfeat{upos\_PROPN} (.052);
  \syntaxfeat{dep\_nsubj} (.033);
  \rawfeat{char\_per\_tok} (.032) \\

& 31 & .339
& \uposfeat{upos\_PROPN} (.134);
  \verbalfeat{verbal\_root} (.061);
  \syntaxfeat{dep\_nsubj} (.056);
  \uposfeat{upos\_NOUN} (.033);
  \rawfeat{n\_tokens} (.029) \\

\addlinespace[2pt]

\multirow{3}{*}{15T}
& 2 & .270
& \rawfeat{n\_tokens} (.199);
  \verbalfeat{verbal\_root} (.032);
  \rawfeat{n\_sentences} (.019);
  \uposfeat{upos\_PUNCT} (.012);
  \rawfeat{char\_per\_tok} (.010) \\

& 16 & .439
& \uposfeat{upos\_NOUN} (.137);
  \syntaxfeat{dep\_nsubj} (.054);
  \rawfeat{char\_per\_tok} (.039);
  \uposfeat{upos\_PROPN} (.037);
  \verbalfeat{verbal\_root} (.032) \\

& 31 & .311
& \uposfeat{upos\_PROPN} (.148);
  \syntaxfeat{dep\_nsubj} (.055);
  \rawfeat{n\_tokens} (.040);
  \verbalfeat{verbal\_root} (.034);
  \uposfeat{upos\_PRON} (.025) \\

\bottomrule
\end{tabularx}
}
\caption{ESS-estimated pointwise ID regression across seven Apertus checkpoints. Scores are mean five-fold cross-validated \(R^2\) values for random forest regression. For each layer, boldface marks the highest score across checkpoints. Parentheses report the mean decrease in held-out \(R^2\) after permutation, averaged over eight shuffles and five cross-validation folds. Background colours indicate the four linguistic feature families: \familykey{rawfamily}{Raw text}, \familykey{uposfamily}{Morphosyntax}, \familykey{syntaxfamily}{Dependency and tree structure}, and \familykey{verbfamily}{Verbal and clausal structure}.}
\label{tab:apertus_checkpoint_ess_importance}
\end{table*}

\subsection{Directionality of linguistic associations with pointwise ID}\label{sec:ale}
Permutation importance shows which features the random forest uses to
predict pointwise ID. To interpret these predictions, we also need to know
whether higher feature values are associated with higher or lower predicted
ID. We examine this question using cross-fitted accumulated local effects
(ALE) \citep{apley2020visualizing}. ALE summarizes how a fitted model's
predictions vary across the observed range of a feature. We estimate these
relationships in all nine model--layer settings, with three layers per
model. The estimation procedure and stability criteria are described in
App.~\ref{app:ale}.

The directions change with depth
(Table~\ref{tab:feature_direction}). At early layers, excerpt length,
measured as the number of tokens, has the largest positive ALE contrast in
all three models. Negative associations at this depth depend on the model.
At intermediate layers, two negative associations recur in every model:
common-noun frequency and the average number of characters per token.
Pronoun frequency is associated with higher predicted ID in Pythia and
Qwen, while nominal-subject frequency has a positive association in Qwen
and Apertus.

At late layers, proper-noun frequency yields the largest positive
ALE contrast in all three models. Nominal subjects and verbal roots also
show positive associations across architectures. Common-noun frequency
remains negatively associated with predicted ID in Pythia and Qwen, where
it is selected among the leading predictors.

These ALE patterns describe relationships learned by the prediction
models. To examine whether similar differences are visible directly in the
data, we also compare the linguistic profiles of excerpts at opposite ends
of each ID distribution. For every model and layer, we use Mann--Whitney
\(U\) tests to compare the 150 excerpts with the lowest ID and the 150 with
the highest ID. The comparisons largely mirror the ALE directions:
low-ID excerpts have higher common-noun frequencies and more characters
per token, whereas high-ID excerpts generally contain more tokens and have
higher frequencies of pronouns, nominal subjects, verbal roots, and proper
nouns.

Together, these recurring differences suggest two broader linguistic
profiles. Lower-ID excerpts are more consistent with technical and
expository writing, combining specialized vocabulary, a higher
concentration of common nouns, and an information-dense style. Higher-ID
excerpts more often resemble informal or narrative writing organized
around participants and events, with shorter words, more pronouns,
explicit subjects, and verbal constructions. Representative excerpts
illustrating these profiles are shown in Table~\ref{tab:mann-whitney} in
App.~\ref{app:ale}.

\subsection{Development across pretraining}\label{par:checkpoints}

The preceding analyses describe linguistic associations with
pointwise ID in the final models. We next ask how these associations change
during pretraining: whether the patterns associated with each depth are
already present at earlier stages, and how they develop thereafter. To
address this question, we repeat the linguistic prediction analysis at
intermediate checkpoints of Apertus and Pythia. At each checkpoint,
we examine both how well linguistic profiles predict pointwise ID, and
which features contribute most to those predictions.

The two checkpoint series cover different parts of pretraining.
Pythia includes stages close to initialization, whereas the earliest
available Apertus checkpoint already follows 210 billion training tokens.
Their trajectories therefore cannot be aligned directly. We interpret
qualitative changes within each model, without treating their checkpoints
as matched stages of development. We focus here on Apertus and report
the Pythia analysis in Appendix~\ref{app:pythia_pretraining}.

Apertus already shows a clear intermediate-layer advantage at its
first available checkpoint
(Table~\ref{tab:apertus_checkpoint_ess_importance}). Linguistic profiles predict pointwise ID most accurately at the intermediate layer, with \(R^2=0.446\). The intermediate layer remains
the most predictable at all seven checkpoints. Prediction scores
fluctuate rather than increasing monotonically, but remain within
relatively narrow ranges at the intermediate and late layers.

The feature rankings show what persists within this overall
pattern. Early-layer predictions rely primarily on excerpt length,
sentence count, and punctuation. At intermediate depth, common-noun
proportion ranks first at every checkpoint, accompanied by proper nouns,
token form, pronouns, and verbal structure. At late depth, proper-noun
proportion consistently ranks first, while nominal subjects and verbal
roots also remain prominent. Across the observed Apertus trajectory,
changes in the prediction models therefore mainly concern the relative
importance of an already established set of predictors. The broad
linguistic roles associated with each depth remain stable.

Pythia complements this picture by tracing the emergence of the
organization that Apertus already displays after substantial pretraining.
Both models show a broadly analogous pattern: surface features dominate
early layers, while nominal and structural properties become more
prominent at intermediate and late depths. Together, the trajectories
suggest that associations between linguistic properties and local
geometry develop relatively early and subsequently persist over the
observed training stages. Their different temporal coverage, however,
does not establish a shared point at which these associations emerge.

\section{Conclusion}\label{sec:conclusion}

A single geometric summary for a dataset cannot show how individual utterances differ within a model's representation space. Estimating pointwise ID against a shared reference set allowed us to examine this variation directly and relate it to several linguistic properties at once. Across Pythia, Qwen, and Apertus, compact linguistic profiles predicted ID at every examined depth, with the strongest relationship consistently appearing in intermediate layers.

The properties associated with this variation changed as representations passed through the models. Early-layer rankings emphasized input length and form, while nominal composition, pronouns, subject dependencies, and verbal structure became more prominent at intermediate and later depths. These associations give a recognizable linguistic character to the geometric differences: excerpts with more common nouns and longer tokens tended to have lower ID, whereas pronouns, nominal subjects, and verbal roots were more often associated with higher ID.

Several analyses support this picture. In the main ESS analysis, linguistic profiles outperformed the matched lexical and log-perplexity baselines, and prediction remained possible after separately removing surface descriptors or frequency controls. Similar depth patterns appeared across the estimators and neighbourhood sizes examined. The checkpoint analyses further traced the emergence of these patterns in Pythia and their persistence over the observed Apertus trajectory. Pointwise ID thus complements aggregate layer profiles by connecting geometric differences between individual representations to observable differences between the text passages that give rise to them.

\section{Limitations}\label{sec:limitations}

Our findings concern short English excerpts from The Pile, three
similarly sized decoder-only models, and representations of the final
non-padding token at selected layers; generalization across languages,
domains, model scales, and representation positions therefore remains open.
Even within this setting, pointwise ID depends on the reference population,
layer, distance metric, neighbourhood size, and estimator. Our checks support
broad patterns across the examined choices, but scores describe local
representation geometry rather than inherent linguistic complexity.

Linguistic interpretation is further constrained by annotation errors
and correlations among features and with topic, genre, and meaning. Baselines
and ablations only partly separate these influences. 

Checkpoint coverage likewise limits developmental conclusions: Pythia
includes checkpoints near initialization, whereas the earliest Apertus
checkpoint follows substantial pretraining. Their unequal coverage cannot
establish a shared point at which linguistic associations emerge; earlier,
denser sampling would support a more direct examination.

\section{Acknowledgments}\label{sec:acknowledgments}
MB and IM received funding from the European Research Council (ERC) under the European Union's Horizon 2020 research and innovation program (grant agreement No. 101019291).

\bibliography{custom}

\appendix

\section{Models, data, and licences}
\label{app:assets}

This appendix lists the principal external assets used in the study,
including pretrained models, datasets, linguistic-analysis resources,
and software packages. For each asset, we provide its public location
and the licence or usage terms stated by its maintainers. Because The
Pile combines sources distributed under different terms, its
component-specific licences apply.
\medskip

\noindent\textbf{Pythia}
\url{https://huggingface.co/EleutherAI/pythia-6.9b-deduped};
licence: Apache-2.0.

\medskip

\noindent\textbf{Qwen}
\url{https://huggingface.co/Qwen/Qwen2.5-7B};
licence: Apache-2.0.

\medskip

\noindent\textbf{Apertus}
\url{https://huggingface.co/swiss-ai/Apertus-8B-2509};
licence: Apache-2.0; usage policy:
\url{https://huggingface.co/swiss-ai/Apertus-8B-2509/blob/main/USAGE_POLICY.md}.

\medskip

\noindent\textbf{The Pile split}
\url{https://huggingface.co/datasets/ArmelR/the-pile-splitted};
The licences and terms of the
underlying Pile components apply:
\url{https://huggingface.co/datasets/EleutherAI/pile}.

\medskip

\noindent\textbf{Profiling-UD}
\url{https://www.ilc.cnr.it/profiling-ud/};

\medskip

\noindent\textbf{scikit-dimension}
\url{https://github.com/scikit-learn-contrib/scikit-dimension};
licence: BSD-3-Clause.

\medskip

\noindent\textbf{pyFCI}
\url{https://github.com/vittorioerba/pyFCI};
licence: MIT.

\medskip

\noindent\textbf{PyTorch}
\url{https://github.com/pytorch/pytorch};
licence: BSD-3-Clause.

\medskip

\noindent\textbf{Transformers}
\url{https://github.com/huggingface/transformers};
licence: Apache-2.0.

\medskip

\noindent\textbf{Datasets}
\url{https://github.com/huggingface/datasets};
licence: Apache-2.0.

\medskip

\section{Reference-set size}
\label{app:reference_size}

Pointwise ID is estimated from the \(k\) nearest representations available in the reference set. Its size therefore determines how densely the surrounding representation space is sampled. If the reference set is too small, the estimated neighbourhood may be sensitive to the particular reference points included, especially for large \(k\), where a substantial portion of the set is used. We therefore examine whether the estimates stabilize as \(n_{\mathrm{ref}}\) increases.

\begin{figure}[htbp]
    \centering
    \includegraphics[width=\linewidth]
    {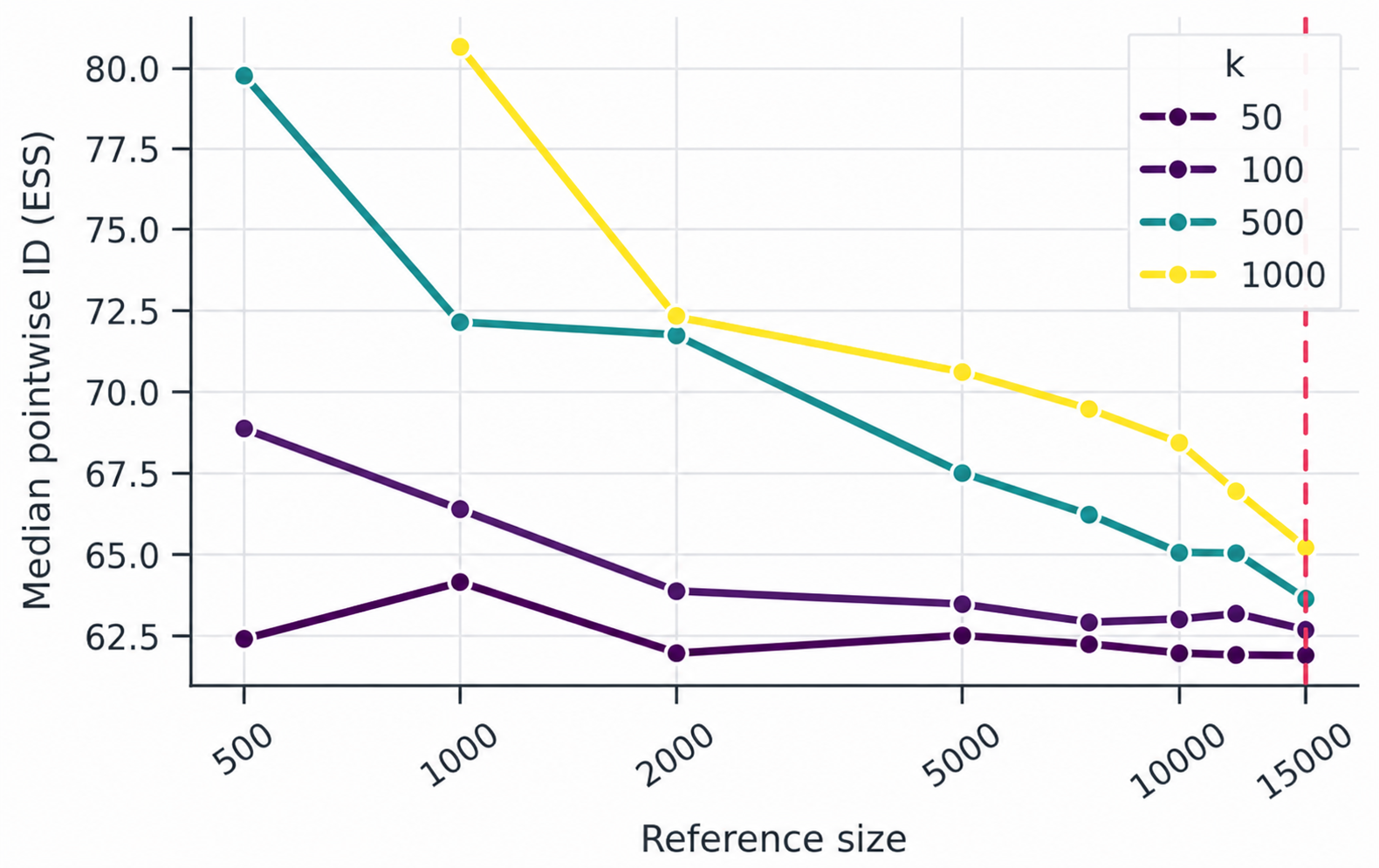}
    \caption{Sensitivity of ESS-estimated pointwise ID to reference-set size for Pythia at layer 16. Each curve shows the median estimate obtained with a different neighbourhood size \(k\).}
    \label{fig:reference_plateau}
\end{figure}

Figure~\ref{fig:reference_plateau} shows substantial changes at small reference sizes, followed by increasingly stable estimates. Between \(n_{\mathrm{ref}}=12{,}000\) and \(15{,}000\), median ID changes by only \(0.03\%\), \(0.80\%\), \(2.14\%\), and \(2.54\%\) for \(k=50,100,500,\) and \(1000\), respectively. The remaining sensitivity is slightly greater for larger neighbourhoods, but still small compared with the changes observed in the lower-reference regime. We therefore use \(n_{\mathrm{ref}}=15{,}000\), the largest value examined, as a conservative operating point that provides stable neighbourhood estimates across the values of \(k\) considered.

\section{Stability across reference sets}
\label{app:stability_reference}
\begin{figure}[t!]
    \centering
    \includegraphics[width=\linewidth]{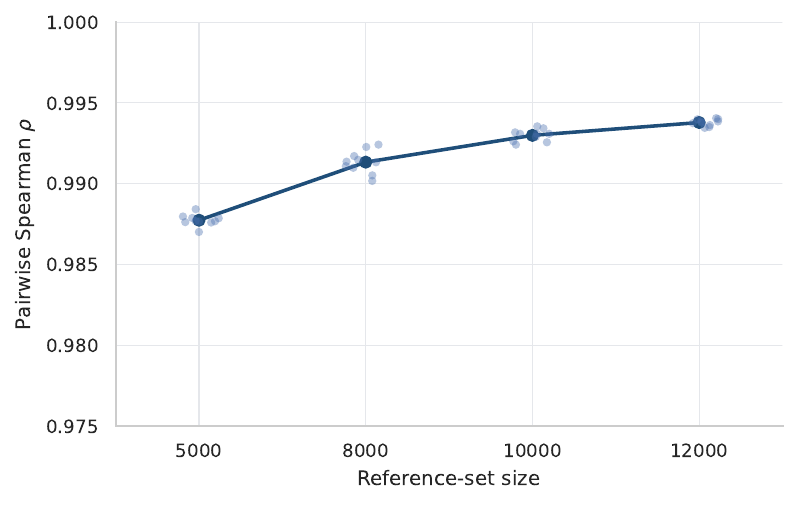}
    \caption{Stability of Pythia-6.9B ESS-estimated pointwise ID rankings under reference-set resampling at layer 16 and \(k=1000\). At each reference-set size, points show the ten pairwise Spearman correlations between rankings obtained from five independently sampled reference subsets; the line shows their mean.}
    \label{fig:ess_reference_stability}
\end{figure}

Because pointwise ID is estimated relative to a sampled reference set, its values could partly depend on which excerpts happen to be included in the sample. This would be problematic if different reference samples assigned substantially different IDs to the same excerpts. To test this possibility, we independently sampled five reference sets at each reference-set size, recomputed the pointwise scores,
and compared the resulting rankings of the evaluation excerpts.
Figure~\ref{fig:ess_reference_stability} shows strong agreement across the five samples: excerpts assigned relatively high or low ID under one reference set generally receive similar rankings under the others. The linguistic analyses therefore do not depend on an unusually favourable reference sample. Instead, the relative pointwise-ID structure is largely preserved when the reference set is resampled.

\section{ID estimators}\label{app:estimators}
\paragraph{Expected Simplex Skewness}
\label{app:pointwise_ess}

Expected Simplex Skewness (ESS) characterizes the angular geometry of
the centred neighbour cloud \citep{johnsson2014low}. We follow the
\texttt{scikit-dimension} implementation \texttt{ESS(ver='a', d=1)} and exclude the target representation from the cloud. We use all distinct neighbour pairs when at most \(5{,}000\)
are available and otherwise sample \(5{,}000\) pairs. The empirical statistic is converted to an ID estimate by interpolation along the analytical reference curve described by \citet{johnsson2014low}.

\paragraph{Maximum-likelihood estimator}
\label{app:pointwise_mle}

The maximum-likelihood estimator (MLE) describes how quickly the
neighbourhood around a target point expands as its radius increases. Let
\[
T_1(x)\leq \cdots \leq T_k(x)
\]
denote the ordered Euclidean distances from query point \(x\) to its \(k\)
nearest reference neighbours. Following \citet{levina2004maximum}, the local
estimate is
\[
\widehat{d}_{\mathrm{MLE}}(x)
=
\left[
\frac{1}{k-1}
\sum_{j=1}^{k-1}
\log\frac{T_k(x)}{T_j(x)}
\right]^{-1}.
\]
The \(k\)-th-neighbour distance defines the outer radius of the local region, while the remaining distances describe how points accumulate within it. When the distances are concentrated near this outer radius, the estimate is higher; when they are distributed over a wider radial range, the estimate is lower.

\paragraph{Full Correlation Integral}
\label{app:pointwise_fci}

The Full Correlation Integral (FCI) was introduced as a global point-cloud estimator and subsequently extended to local neighbourhoods \citep{erba2019intrinsic}. For each target representation, we construct a cloud containing the target and its \(k\) nearest reference neighbours, centre it, and project its points onto the unit sphere. We then fit the finite-radius correlation-integral curve using \texttt{pyFCI.local\_FCI}. We pass \(k+1\) as the cloud size because
the implementation counts the target itself, sample at most \(500\) point pairs with replacement, and use at most \(500\) points in the curve fit. Query-specific random seeds are derived deterministically from a base seed of \(42\). The reported estimate restores the radial
degree of freedom removed by spherical normalization. Non-positive or non-finite estimates, and fits whose estimated radius differs from one by more than \(0.1\), are treated as missing.

\paragraph{Local estimates and dataset-level summaries}
Let
\[
\mathcal{N}_{\ell,k}(x)
=
\{
\mathbf{z}_1(x),\ldots,\mathbf{z}_k(x)
\}
\]
denote the \(k\) nearest representations to the query point
\(\mathbf{h}_{\ell}(x)\) in the fixed reference set
\(\mathcal{R}_{\ell}\), ordered by their Euclidean distances
\begin{align*}
& T_i(x) = \left\| \mathbf{h}_{\ell}(x)-\mathbf{z}_i(x) \right\|_2, \\
& T_1(x) \leq\cdots\leq T_k(x).
\end{align*}
Throughout this paragraph, we suppress the dependence on the fixed
reference set and its size \(n_{\mathrm{ref}}\) in order to simplify
the notation.

Pointwise ID assigns a local geometric estimate to the query \(x\)
through the neighbourhood that \(x\) selects. For MLE and ESS, this
target-specific quantity is part of the estimator's original local
formulation \citep{levina2004maximum,johnsson2014low}. When a single
dataset-level value is required, the local estimates can subsequently
be aggregated, for example as
\begin{align*}
& \widehat d^{\,\mathrm{agg}}_{A}(\ell,k)
=
\operatorname{Agg}_{x\in\mathcal{E}}
\widehat d_{A}(x;\ell,k),
\\
&A\in\{\mathrm{MLE},\mathrm{ESS}\},
\end{align*}

where \(\mathcal{E}\) is the evaluation set and \(\operatorname{Agg}\) may be a mean or median. 

FCI differs in that its basic formulation estimates the dimension of a complete point cloud. Its multiscale extension applies the same estimator to a neighbourhood surrounding a selected point, with the scale controlled either by a cutoff radius or by the number of nearest neighbours \citep{erba2019intrinsic}. We use the latter construction to obtain one FCI estimate for each target point at a fixed \(k\).

\section{Choosing \(k\)}\label{app:k}
\begin{figure}[tbp]
    \centering
    \includegraphics[width=1\linewidth]{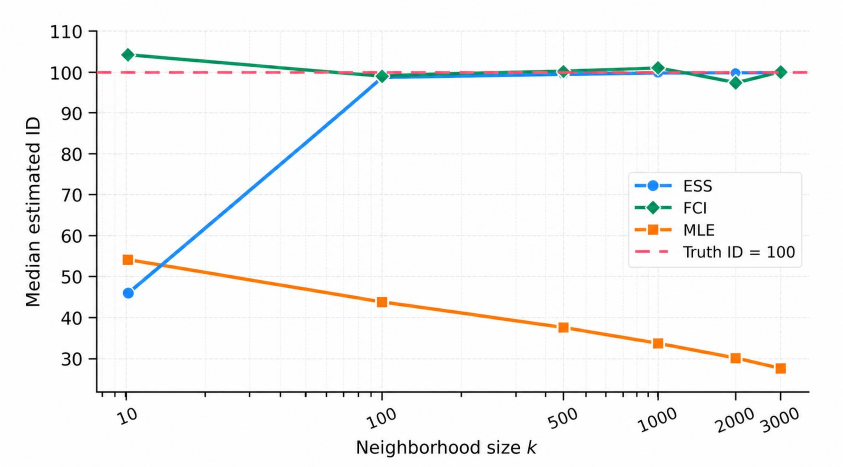}
    \caption{Median ID estimates from ESS, FCI, and MLE on a synthetic Gaussian cloud with true intrinsic dimension \(D=100\) and \(n_{\mathrm{ref}}=15{,}000\), across neighbourhood sizes. The dashed horizontal line marks the true dimension.}
    \label{fig:estimators}
\end{figure}

As a calibration check, we evaluated our three estimators on synthetic Gaussian-cloud data with known dimension $D=100$, using $n_{\mathrm{ref}}=15000$ and several neighbourhood sizes. Figure~\ref{fig:estimators} shows that very small neighbourhoods are unstable. At very small neighbourhood sizes, ESS underestimates the target dimension, while FCI overestimates it. MLE displays substantial negative bias under this synthetic configuration. Since our main experiments use $k=1000$, this places them well outside the visibly unstable small-$k$ regime.

\begin{figure}[htbp]
    \centering
    \includegraphics[width=\linewidth]{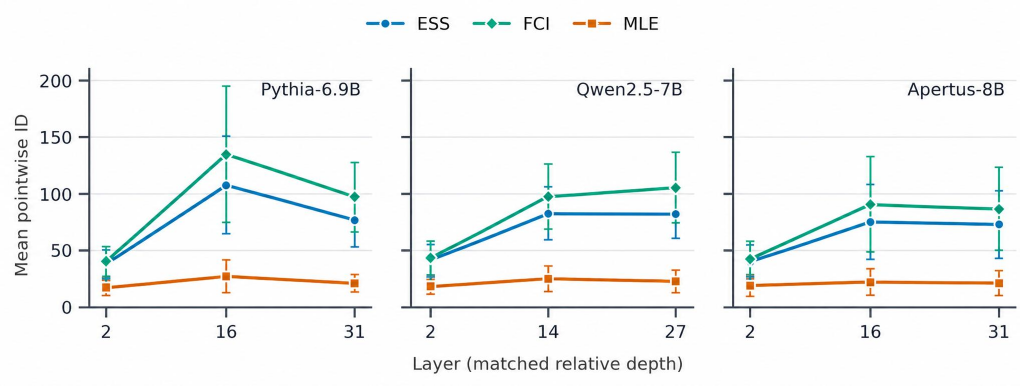}
    \caption{Mean pointwise ID (\(\pm1\sigma\), where \(\sigma\) is the standard deviation across excerpts) estimated with ESS, FCI, and MLE across approximately matched relative depths in Pythia, Qwen, and Apertus.}
    \label{fig:scores_on_layers}
\end{figure}

\paragraph{Complementarity of the estimators}
The three estimators use different aspects of neighbourhood geometry, so they are not expected to return the same ID values. MLE relies on distances from the target to its neighbours, ESS characterizes the angular geometry of the centred neighbour cloud, and FCI uses pairwise distances in the centered, spherically normalized cloud containing the target and its neighbours. Robustness therefore means that the estimators recover similar patterns across layers and linguistic properties, not that they produce identical numerical estimates.
Figure~\ref{fig:scores_on_layers} compares the aggregate depth profiles recovered by the three estimators. ESS and FCI both show an increase from early to intermediate depth. ESS reaches an intermediate-layer maximum or plateau in all three models, whereas the highest mean FCI estimate occurs at the late layer in Qwen. MLE produces lower ID estimates overall and a flatter depth profile, making this peak less pronounced. This is consistent with its downward bias in the calibration analysis.

\section{Hyperparameters and cross-validation protocol}
\label{app:hyp}

All model hyperparameters were fixed before evaluation and were not tuned on held-out observations. Unless otherwise stated, we used five-fold cross-validation with random seed 42. For each architecture, layer, and target,
the same fold assignments were reused across algorithms and feature-set comparisons, allowing paired comparisons on identical held-out examples.

\paragraph{Profiling-UD models}

Random forest regressors and classifiers used 300 trees, a
minimum leaf size of two observations, and random seed 42. The
classification forests balanced class weights within each bootstrap
sample. Missing predictor values were imputed using medians estimated on the training fold.

\paragraph{Permutation importance}

Permutation importance was computed exclusively on held-out folds. For
each fitted forest, every predictor was independently shuffled eight
times within the validation fold, and importance was measured as the
resulting decrease in \(R^2\) for regression or ROC-AUC for
classification. Values were averaged across permutations and folds. We
retained features with positive mean importance whose fold-level
importance was positive in at least four of the five folds.

\section{Robustness of predictive performance to the choice of ID estimator}
\label{app:id_estimator_robustness}

The main analysis uses ESS as the pointwise-ID estimation method. To determine
whether the results depend on this choice, we repeated the regression,
feature-ablation, and held-out permutation-importance analyses using MLE
and FCI. We retained the same excerpts, 46 predictors,
layers, \(n_{\mathrm{ref}}=15{,}000\), \(k=1000\), and five-fold
cross-validation protocol. Table~\ref{tab:alternative-estimator-ablation} reports predictive performance and ablation controls, while Table~\ref{tab:alternative-estimator-importance} reports the five
largest held-out permutation importances.

\begin{table}[t]
\centering
\footnotesize
\setlength{\tabcolsep}{3.5pt}
\renewcommand{\arraystretch}{1.06}
\renewcommand{\tabularxcolumn}[1]{m{#1}}

\begin{tabularx}{\columnwidth}{l c c
  >{\centering\arraybackslash}X !{\hspace{2pt}}
  >{\centering\arraybackslash}X}
\toprule
\textbf{Model} & \textbf{Layer} & \textbf{Full} &
\textbf{No\newline frequency} & \textbf{No raw\newline features} \\
\midrule

\rowcolor{black!7}
\multicolumn{5}{l}{\textbf{FCI}} \\

\multirow{3}{*}{Pythia-6.9B}
& 2  & .259 & \dnone{.257} & \dmedium{.213} \\
& 16 & .424 & \dnone{.423} & \dmedium{.396} \\
& 31 & .208 & \dnone{.207} & \dsmall{.201} \\

\addlinespace[2pt]

\multirow{3}{*}{Qwen2.5-7B}
& 2  & .204 & \dnone{.203} & \dmedium{.168} \\
& 14 & .482 & \dnone{.481} & \dsmall{.464} \\
& 27 & .359 & \dnone{.358} & \dsmall{.344} \\

\addlinespace[2pt]

\multirow{3}{*}{Apertus-8B}
& 2  & .253 & \improved{.253} & \dlarge{.194} \\
& 16 & .421 & \improved{.421} & \dsmall{.404} \\
& 31 & .286 & \dnone{.284} & \dsmall{.266} \\

\addlinespace[3pt]
\midrule

\rowcolor{black!7}
\multicolumn{5}{l}{\textbf{MLE}} \\

\multirow{3}{*}{Pythia-6.9B}
& 2  & .127 & \improved{.128} & \dnone{.122} \\
& 16 & .259 & \improved{.260} & \dmedium{.224} \\
& 31 & .213 & \dnone{.212} & \dsmall{.194} \\

\addlinespace[2pt]

\multirow{3}{*}{Qwen2.5-7B}
& 2  & .049 & \dnone{.049} & \dsmall{.033} \\
& 14 & .239 & \dnone{.239} & \dsmall{.228} \\
& 27 & .159 & \dsmall{.153} & \dsmall{.144} \\

\addlinespace[2pt]

\multirow{3}{*}{Apertus-8B}
& 2  & .079 & \improved{.080} & \dsmall{.062} \\
& 16 & .268 & \dnone{.267} & \dmedium{.247} \\
& 31 & .101 & \dnone{.097} & \dsmall{.088} \\

\bottomrule
\end{tabularx}

\caption{Pointwise ID regression and feature-ablation controls for FCI and MLE. Values are mean five-fold cross-validated \(R^2\) scores for random forest regression, rounded to three decimal places. \textit{No frequency} removes the three frequency controls, while \textit{No raw features} removes the four surface descriptors. In the ablation columns, shading represents \(\Delta R^2=R^2_{\mathrm{Full}}-R^2_{\mathrm{Ablated}}\): grey for \(0\leq\Delta R^2<.005\), yellow for \(.005\leq\Delta R^2<.020\), orange for \(.020\leq\Delta R^2<.050\), red for \(.050\leq\Delta R^2<.100\), and bright red for \(\Delta R^2\geq.100\). Blue indicates improvement after ablation.}
\label{tab:alternative-estimator-ablation}
\end{table}

\begin{table*}[!tp]
\centering
\footnotesize
\setlength{\fboxsep}{0.5pt}

\begin{tabularx}{\textwidth}{
    @{}
    c@{\hspace{4pt}}
    >{\raggedright\arraybackslash}X
    @{}
}
\toprule
\textbf{Layer} &
\textbf{Five largest held-out RF permutation importances
\({(\Delta R^2)}\)} \\
\midrule
\multicolumn{2}{c}{\textbf{FCI}} \\
\rowcolor{black!5}

\multicolumn{2}{@{}l@{}}{\textbf{Pythia-6.9B}} \\

2 &
\rawfeature{n\_tokens} (.128),
\uposfeature{upos\_PROPN} (.075),
\depfeature{dep\_nsubj} (.024),
\rawfeature{char\_per\_tok} (.020),
\uposfeature{upos\_PRON} (.017) \\

16 &
\uposfeature{upos\_NOUN} (.143),
\rawfeature{char\_per\_tok} (.079),
\uposfeature{upos\_PROPN} (.066),
\verbalfeature{verbal\_root} (.042),
\uposfeature{upos\_PRON} (.035) \\

31 &
\uposfeature{upos\_PROPN} (.093),
\uposfeature{upos\_NOUN} (.071),
\verbalfeature{verbal\_root} (.034),
\uposfeature{upos\_PRON} (.022),
\uposfeature{upos\_ADV} (.018) \\

\addlinespace[2pt]
\rowcolor{black!5}

\multicolumn{2}{@{}l@{}}{\textbf{Qwen2.5-7B}} \\

2 &
\rawfeature{n\_tokens} (.128),
\verbalfeature{verbal\_root} (.053),
\uposfeature{upos\_PUNCT} (.019),
\rawfeature{char\_per\_tok} (.018),
\uposfeature{upos\_NUM} (.015) \\

14 &
\rawfeature{char\_per\_tok} (.152),
\uposfeature{upos\_PRON} (.065),
\uposfeature{upos\_PROPN} (.058),
\depfeature{dep\_nsubj} (.047),
\uposfeature{upos\_NOUN} (.035) \\

27 &
\uposfeature{upos\_PROPN} (.154),
\verbalfeature{verbal\_root} (.139),
\uposfeature{upos\_NOUN} (.062),
\rawfeature{char\_per\_tok} (.031),
\uposfeature{upos\_ADJ} (.015) \\

\addlinespace[2pt]
\rowcolor{black!5}

\multicolumn{2}{@{}l@{}}{\textbf{Apertus-8B}} \\

2 &
\rawfeature{n\_tokens} (.215),
\verbalfeature{verbal\_root} (.032),
\rawfeature{n\_sentences} (.012),
\uposfeature{upos\_PUNCT} (.010),
\rawfeature{char\_per\_tok} (.008) \\

16 &
\uposfeature{upos\_NOUN} (.113),
\uposfeature{upos\_PROPN} (.047),
\verbalfeature{verbal\_root} (.044),
\depfeature{dep\_nsubj} (.041),
\rawfeature{char\_per\_tok} (.041) \\

31 &
\uposfeature{upos\_PROPN} (.188),
\depfeature{dep\_nsubj} (.051),
\verbalfeature{verbal\_root} (.034),
\rawfeature{n\_tokens} (.033),
\uposfeature{upos\_PRON} (.030) \\

\addlinespace[3pt]
\midrule
\multicolumn{2}{c}{\textbf{MLE}} \\
\rowcolor{black!5}

\multicolumn{2}{@{}l@{}}{\textbf{Pythia-6.9B}} \\

2 &
\uposfeature{upos\_PROPN} (.091),
\uposfeature{upos\_PUNCT} (.040),
\uposfeature{upos\_NOUN} (.026),
\depfeature{dep\_nsubj} (.022),
\uposfeature{upos\_NUM} (.017) \\

16 &
\uposfeature{upos\_NOUN} (.083),
\depfeature{subj\_pre} (.070),
\rawfeature{char\_per\_tok} (.058),
\uposfeature{upos\_PROPN} (.023),
\depfeature{dep\_nsubj} (.014) \\

31 &
\uposfeature{upos\_PROPN} (.079),
\rawfeature{char\_per\_tok} (.043),
\depfeature{dep\_nsubj} (.033),
\uposfeature{upos\_ADJ} (.025),
\verbalfeature{verbal\_root} (.025) \\

\addlinespace[2pt]
\rowcolor{black!5}

\multicolumn{2}{@{}l@{}}{\textbf{Qwen2.5-7B}} \\

2 &
\rawfeature{n\_tokens} (.041),
\uposfeature{upos\_PROPN} (.027),
\verbalfeature{verbal\_root} (.022),
\rawfeature{char\_per\_tok} (.019),
\uposfeature{upos\_ADJ} (.011) \\

14 &
\rawfeature{char\_per\_tok} (.092),
\depfeature{dep\_nsubj} (.038),
\uposfeature{upos\_PROPN} (.029),
\depfeature{dep\_compound} (.021),
\uposfeature{upos\_NOUN} (.018) \\

27 &
\uposfeature{upos\_PROPN} (.127),
\rawfeature{char\_per\_tok} (.034),
\depfeature{dep\_nsubj} (.024),
\verbalfeature{verbal\_root} (.018),
\uposfeature{upos\_PRON} (.017) \\

\addlinespace[2pt]
\rowcolor{black!5}

\multicolumn{2}{@{}l@{}}{\textbf{Apertus-8B}} \\

2 &
\rawfeature{n\_tokens} (.086),
\uposfeature{upos\_PROPN} (.066),
\uposfeature{upos\_NOUN} (.039),
\uposfeature{upos\_PUNCT} (.020),
\uposfeature{upos\_NUM} (.017) \\

16 & 
\uposfeature{upos\_NOUN} (.048), \rawfeature{char\_per\_tok} (.043), \depfeature{dep\_nsubj} (.033), \uposfeature{upos\_PROPN} (.025), \rawfeature{n\_tokens} (.021) \\

31 &
\uposfeature{upos\_PROPN} (.124),
\rawfeature{char\_per\_tok} (.032),
\depfeature{dep\_nsubj} (.027),
\freqfeature{median\_frequency\_class} (.021),
\rawfeature{n\_tokens} (.020) \\

\bottomrule
\end{tabularx}

\smallskip
\caption{Most predictive features for FCI- and MLE-estimated pointwise ID regression. Parentheses report the mean decrease in held-out \(R^2\) after permutation, averaged over eight shuffles and five cross-validation folds. Background colours indicate the linguistic feature families and frequency controls: \familykey{rawfamily}{Raw text}, \familykey{uposfamily}{Morphosyntax}, \familykey{syntaxfamily}{Dependency and tree structure}, \familykey{verbfamily}{Verbal and clausal structure}, and \familykey{frequencyfamily}{Lexical frequency}.}
\label{tab:alternative-estimator-importance}
\end{table*}

The alternative estimators reproduce the main depth pattern. For both
MLE and FCI, random forest performance peaks at the intermediate layer
in every model. Removing frequency summaries changes performance by less
than \(0.009\ R^2\), while removing raw-text descriptors generally
produces a larger reduction without changing the layer ordering.

The leading predictors also overlap substantially with the ESS
analysis. Measures of input extent and nominal composition recur at early depth, while noun and proper-noun proportions, average token length, nominal subjects, and verbal roots become prominent at intermediate and late depth. A frequency summary appears in only one top-five ranking, for Apertus MLE at the late layer. Overall, the main conclusions are therefore not specific to ESS.
\section{Robustness of predictive performance to layer choice and
neighbourhood size}
\label{app:extra_analysis}

The main experiments represent early, intermediate, and late depth using
one layer from each region and \(k=1000\). We test whether the resulting
depth pattern depends on either choice. For Qwen and Pythia, we repeat the
random forest regression across neighbourhood sizes from \(k=10\) to
\(k=3000\), including an adjacent layer at each depth, and conduct the
analysis separately for MLE, ESS, and FCI. Because they exhibited similar
patterns, we expected the conclusions to extend to Apertus and did not
repeat the full grid for that model.
\begin{table}[t!]
\centering
\scriptsize
\setlength{\tabcolsep}{5pt}
\renewcommand{\arraystretch}{0.94}
\begin{tabular}{@{}lcrrrrr@{}}
\toprule
\textbf{Target} & \textbf{Layer} & \multicolumn{5}{c}{\textbf{Neighbourhood size $k$}} \\
\cmidrule(lr){3-7}
& & \textbf{10} & \textbf{100} & \textbf{500} & \textbf{2000} & \textbf{3000} \\
\midrule
\multicolumn{7}{@{}l}{\textbf{Pythia-6.9B}} \\
\addlinespace[1pt]
ESS & 1 & .142 & .185 & .200 & .258 & .299 \\
 & 2 & .167 & .184 & .204 & .374 & .491 \\
 & 16 & .171 & .327 & .457 & .507 & .515 \\
 & 17 & .159 & .312 & .438 & .502 & .497 \\
 & 31 & .076 & .111 & .230 & .248 & .244 \\
 & 32 & .097 & .234 & .351 & .456 & .506 \\
\addlinespace[2pt]
FCI & 1 & .056 & .179 & .184 & .235 & .347 \\
 & 2 & .059 & .163 & .183 & .381 & .495 \\
 & 16 & .068 & .299 & .397 & .417 & .434 \\
 & 17 & .064 & .295 & .395 & .390 & .421 \\
 & 31 & .017 & .043 & .198 & .214 & .245 \\
 & 32 & .036 & .205 & .334 & .455 & .503 \\
 \addlinespace[2pt]
 MLE & 1 & .024 & .087 & .062 & .069 & .071 \\
 & 2 & .022 & .071 & .099 & .148 & .147 \\
 & 16 & .058 & .168 & .247 & .268 & .265 \\
 & 17 & .055 & .165 & .239 & .254 & .255 \\
 & 31 & -.003 & .088 & .195 & .205 & .176 \\
 & 32 & .000 & .051 & .136 & .230 & .229 \\
 \midrule
\multicolumn{7}{@{}l}{\textbf{Qwen2.5-7B}} \\
\addlinespace[1pt]
ESS & 1 & .112 & .150 & .146 & .165 & .173 \\
 & 2 & .124 & .219 & .210 & .323 & .399 \\
 & 14 & .173 & .325 & .451 & .582 & .609 \\
 & 15 & .178 & .322 & .426 & .573 & .612 \\
 & 27 & .136 & .249 & .396 & .470 & .523 \\
 & 28 & .073 & .160 & .200 & .378 & .410 \\
\addlinespace[2pt]
FCI & 1 & .019 & .142 & .129 & .179 & .192 \\
 & 2 & .009 & .189 & .176 & .320 & .365 \\
 & 14 & .065 & .294 & .407 & .496 & .510 \\
 & 15 & .075 & .291 & .389 & .483 & .474 \\
 & 27 & .048 & .202 & .314 & .407 & .435 \\
 & 28 & .002 & .137 & .180 & .359 & .383 \\
 \addlinespace[2pt]
MLE & 1 & .021 & .070 & .031 & -.002 & -.005 \\
 & 2 & .036 & .093 & .051 & .058 & .068 \\
 & 14 & .056 & .188 & .228 & .271 & .292 \\
 & 15 & .070 & .165 & .222 & .258 & .276 \\
 & 27 & .009 & .086 & .133 & .178 & .164 \\
 & 28 & -.011 & .048 & .128 & .270 & .282 \\
\addlinespace[2pt]
\bottomrule
\end{tabular}
\caption{Sensitivity of Profiling-UD regression to neighbourhood size and layer in Pythia and Qwen. Values are mean five-fold cross-validated \(R^2\) scores for random forest regression using the same 46 predictors as the main analysis. ESS results at \(k=1000\) are reported in Table~\ref{tab:main-results}; FCI and MLE results at \(k=1000\) are reported in Table~\ref{tab:alternative-estimator-ablation}.}
\label{tab:k-layer-sensitivity}
\end{table}

\begin{table*}[!tp]
\centering
\scriptsize
\setlength{\tabcolsep}{4pt}
\renewcommand{\arraystretch}{1.16}

\begin{tabularx}{\textwidth}{
    @{}
    ll
    >{\raggedright\arraybackslash}X
    >{\raggedright\arraybackslash}X
    cc
    @{}
}
\toprule
\textbf{Model} &
\textbf{Depth} &
\textbf{Main top three} &
\textbf{Main-feature recurrence} &
\shortstack{\textbf{\(k\)-}\\\textbf{overlap}} &
\shortstack{\textbf{Adjacent}\\\textbf{overlap}} \\
\midrule

Pythia-6.9B
& Early (2/1)
& \rawfeat{n\_tokens} (.159);
  \uposfeat{upos\_PROPN} (.041);
  \uposfeat{upos\_PUNCT} (.033)
& \rawfeat{n\_tokens} (\(5/5\));
  \uposfeat{upos\_PROPN} (\(5/5\));
  \uposfeat{upos\_PUNCT} (\(3/5\))
& \(3/5\) & \(4/5\) \\

& Middle (16/17)
& \uposfeat{upos\_NOUN} (.149);
  \rawfeat{char\_per\_tok} (.091);
  \uposfeat{upos\_PRON} (.074)
& \uposfeat{upos\_NOUN} (\(5/5\));
  \rawfeat{char\_per\_tok} (\(4/5\));
  \uposfeat{upos\_PRON} (\(5/5\))
& \(4/5\) & \(5/5\) \\

& Late (31/32)
& \uposfeat{upos\_NOUN} (.075);
  \uposfeat{upos\_PROPN} (.061);
  \verbfeat{verbal\_root} (.040)
& \uposfeat{upos\_NOUN} (\(5/5\));
  \uposfeat{upos\_PROPN} (\(3/5\));
  \verbfeat{verbal\_root} (\(2/5\))
& \(3/5\) & \(3/5\) \\

\addlinespace[3pt]
\midrule
\addlinespace[3pt]

Qwen2.5-7B
& Early (2/1)
& \rawfeat{n\_tokens} (.163);
  \verbfeat{verbal\_root} (.042);
  \uposfeat{upos\_PUNCT} (.033)
& \rawfeat{n\_tokens} (\(5/5\));
  \verbfeat{verbal\_root} (\(3/5\));
  \uposfeat{upos\_PUNCT} (\(3/5\))
& \(3/5\) & \(3/5\) \\

& Middle (14/15)
& \rawfeat{char\_per\_tok} (.136);
  \uposfeat{upos\_PRON} (.107);
  \synfeat{dep\_nsubj} (.053)
& \rawfeat{char\_per\_tok} (\(4/5\));
  \uposfeat{upos\_PRON} (\(5/5\));
  \synfeat{dep\_nsubj} (\(3/5\))
& \(4/5\) & \(5/5\) \\

& Late (27/28)
& \verbfeat{verbal\_root} (.133);
  \uposfeat{upos\_NOUN} (.125);
  \uposfeat{upos\_PROPN} (.123)
& \verbfeat{verbal\_root} (\(4/5\));
  \uposfeat{upos\_NOUN} (\(5/5\));
  \uposfeat{upos\_PROPN} (\(3/5\))
& \(4/5\) & \(3/5\) \\
\bottomrule
\end{tabularx}

\caption{Stability of held-out permutation-importance rankings for ESS-estimated pointwise ID. \textit{Main top three} lists the leading features at \(k=1000\), with mean \(\Delta R^2\) in parentheses; \textit{recurrence} counts their top-five appearances across the five alternative \(k\) values. The final columns report median top-five overlap with the main ranking across alternative \(k\) values and top-five overlap with the adjacent layer at \(k=1000\), respectively. Background colours indicate the four linguistic feature families: \familykey{rawfamily}{Raw text}, \familykey{uposfamily}{Morphosyntax}, \familykey{syntaxfamily}{Dependency and tree structure}, and \familykey{verbfamily}{Verbal and clausal structure}.}
\label{tab:importance-sensitivity}
\end{table*}

Two patterns are consistent across the models (Table~\ref{tab:k-layer-sensitivity}). First, \(k=10\) generally produces weaker and more variable prediction, suggesting that very small neighbourhoods provide an unstable description of local geometry. Performance usually increases with \(k\), particularly for ESS and FCI, as more neighbours make the estimates more stable. 

Second, the intermediate layers were generally the most predictable across neighbourhood sizes and estimators. The similar results obtained for adjacent intermediate layers
show that this pattern is not an artefact of selecting one particular
layer. Some late-layer Pythia results approach or exceed the intermediate
layers for FCI at the largest neighbourhoods, indicating that the precise
layer ranking can depend on the geometric scale. 

Finally, we test whether robustness in predictive performance is
accompanied by stability in linguistic interpretation.
Table~\ref{tab:importance-sensitivity} compares the main ESS
permutation-importance ranking, computed at \(k=1000\), with rankings
obtained using alternative neighbourhood sizes and an adjacent layer.
The first layer listed for each depth is the main layer and the second
is the adjacent layer. \textit{Main top three} reports the leading
features in the reference ranking, while \textit{main-feature
recurrence} counts how often each feature remains among the top five
across the five alternative \(k\) values. \textit{\(k\)-overlap} instead
considers the complete top-five sets and reports their median overlap
with the reference ranking; \textit{adjacent overlap} compares the
complete top-five sets of the two layers at \(k=1000\). For example,
\texttt{n\_tokens} (\(5/5\)) for Pythia's early layer means that it
appears among the top five at every alternative \(k\), whereas a
\(k\)-overlap of \(3/5\) means that the alternative rankings share a
median of three features with the reference top five.

The rankings are most stable at intermediate depth, where both models
retain a median of four of the five reference features across
neighbourhood sizes and all five at the adjacent layer. Early and late
layers exhibit greater variation while preserving the same broad
linguistic feature families.

\section{Robustness across The Pile training and test splits}
\label{app:extra_results}
We tested whether the pointwise-ID results depended on the Pile partition
from which the evaluation excerpts were drawn. Both sets were constructed
using the same preprocessing procedure: source excerpts were normalized,
filtered to retain prose, and truncated at the last complete
punctuation boundary within 100 Pythia-tokenizer tokens. The 15,000 reference excerpts
were selected first from a Pile training shard. The next 2,000
excerpts were chosen at random and formed the \texttt{main\_set}, which is disjoint from the reference
set and is the evaluation set used in all other analyses in this paper.
For this robustness check, we constructed \texttt{extra\_set} from 2,000
eligible excerpts in the Pile test split, applying the same preprocessing
and selection criteria. The two evaluation sets were equal in size and had
closely matched distributions across the constituent Pile sources.

This comparison addresses a potential effect of corpus-partition
provenance. Pythia was pretrained on data derived from the Pile training
split, and some of the same material may also have occurred in Qwen's
pretraining corpus. This analysis should therefore be understood as a
robustness check across Pile partitions. For Pythia, using \(k=1000\) and \(n_{\mathrm{ref}}=15{,}000\), the means and standard deviations of MLE,
ESS, and FCI were closely aligned between \texttt{main\_set} and
\texttt{extra\_set} at all three analysed layers
(Table~\ref{tab:pile_split_id_robustness}). We additionally tested whether
the sets could be distinguished from their joint ID profiles using all
nine scores (three estimators at three layers). Five-fold cross-validation
assigned identical normalized excerpts to the same fold. Classification
remained at chance level, providing no evidence of a systematic difference
between the pointwise-ID profiles of the two sets.
\begin{table}[t]
\centering
\scriptsize
\setlength{\tabcolsep}{2.5pt}
\renewcommand{\arraystretch}{1.08}

\begin{tabular*}{\columnwidth}{
    @{\extracolsep{\fill}}
    ccrrc
    @{}
}
\toprule
\textbf{Layer}
& \textbf{Est.}
& \textbf{\texttt{main}}
& \textbf{\texttt{extra}}
& \({|d|}\) \\
\midrule
\multirow{3}{*}{2}
& ESS & \(38.300 \pm 12.610\) & \(38.250 \pm 12.720\) & .004 \\
& FCI & \(40.600 \pm 13.270\) & \(40.250 \pm 13.430\) & .026 \\
& MLE & \(17.460 \pm 7.000\)  & \(17.120 \pm 7.110\)  & .049 \\
\addlinespace[2pt]
\multirow{3}{*}{16}
& ESS & \(108.110 \pm 42.840\) & \(107.750 \pm 42.660\) & .008 \\
& FCI & \(135.290 \pm 60.210\) & \(134.990 \pm 59.700\) & .005 \\
& MLE & \(27.410 \pm 14.700\)   & \(27.220 \pm 14.290\)   & .014 \\
\addlinespace[2pt]
\multirow{3}{*}{31}
& ESS & \(77.060 \pm 23.800\) & \(76.820 \pm 23.650\) & .010 \\
& FCI & \(97.270 \pm 30.640\) & \(97.200 \pm 30.640\) & .002 \\
& MLE & \(21.290 \pm 7.610\)  & \(21.230 \pm 7.820\)  & .007 \\
\bottomrule
\end{tabular*}
\caption{Pointwise ID distributions across the Pile evaluation splits for Pythia, using \(k=1000\) and \(n_{\mathrm{ref}}=15{,}000\). Columns \texttt{main} and \texttt{extra} refer to \texttt{main\_set} and \texttt{extra\_set}, respectively. Values are means \(\pm\) standard deviations across excerpts; \(|d|\) is the absolute standardized mean difference based on the pooled standard deviation. All \(|d|\) values are below \(.050\).}
\label{tab:pile_split_id_robustness}
\end{table}

\section{Binary ID classification} \label{app:classification}
Continuous regression tests whether linguistic profiles explain variation
across the complete ID distribution. As a complementary and deliberately
simpler task, we examined whether the same profiles distinguish excerpts at
its two extremes. This comparison also makes it easier to relate recurring
feature-importance patterns to broader linguistic profiles.

For each model and layer, excerpts at or below the 20th percentile of the post-merge ESS-estimated ID distribution were assigned to the low-ID class, while those at or above the 80th percentile were assigned to the high-ID class. The intermediate observations were excluded, and ties at the percentile
boundaries were retained. Depending on these ties, the resulting datasets
contained between 798 and 800 excerpts.

We fitted a random forest classifier to the same 46 predictors used in the full regression analysis. The forest used 300 trees, a minimum leaf size of two, and class weights balanced within each bootstrap sample. Performance was evaluated with five-fold stratified cross-validation, using seed 42, and is reported as mean held-out ROC-AUC.

The ESS results in Table~\ref{tab:lexical_classification} show that random forest classifiers distinguish the two groups well in all three models. As in continuous regression, performance is highest at intermediate depth

\paragraph{Classifier feature importance}

For the random forests (Table~\ref{tab:ess-classification-features}), feature importance was measured by held-out
permutation. Within each fold, each predictor was shuffled eight times in
the test portion, and importance was defined as the resulting decrease in
ROC-AUC, averaged across repetitions and folds. 

The classifier rankings closely resemble those obtained for continuous regression. Token count is the leading early-layer predictor in all three models, with sentence count, punctuation, and token length also appearing. At intermediate depth, the rankings shift toward nominal composition, pronouns, subjects, and token length. Late layers emphasize proper nouns in Pythia and Apertus and verbal-root prevalence in Qwen, while common nouns and subject dependencies remain recurrent. None of the three frequency controls appears among the five leading classifier features. Thus, classification supports the same general interpretation as regression: depth reweights a largely shared set of surface and linguistic properties rather than introducing an entirely different predictive profile.

\begin{table*}[t]
\centering
\footnotesize
\setlength{\tabcolsep}{2.5pt}
\renewcommand{\arraystretch}{1.10}
\renewcommand{\tabularxcolumn}[1]{m{#1}}

\begin{tabularx}{\textwidth}{@{}c *{5}{>{\centering\arraybackslash}X}@{}}
\toprule
\textbf{Layer} & \multicolumn{5}{c}{\textbf{Five largest RF permutation importances} \((\Delta\mathrm{AUC})\)} \\
\cmidrule(l){2-6}
& \textbf{1} & \textbf{2} & \textbf{3} & \textbf{4} & \textbf{5} \\
\midrule

\rowcolor{black!5}[0pt][0pt]
\multicolumn{6}{@{}l@{}}{\textbf{Pythia-6.9B}} \\
2
& \mbox{\rawfeat{n\_tokens} (.070)}
& \mbox{\uposfeat{upos\_PROPN} (.015)}
& \mbox{\synfeat{dep\_nsubj} (.015)}
& \mbox{\uposfeat{upos\_PART} (.012)}
& \mbox{\uposfeat{upos\_PRON} (.009)} \\

16
& \mbox{\rawfeat{char\_per\_tok} (.019)}
& \mbox{\uposfeat{upos\_NOUN} (.015)}
& \mbox{\uposfeat{upos\_PROPN} (.009)}
& \mbox{\uposfeat{upos\_PRON} (.005)}
& \mbox{\synfeat{dep\_nsubj} (.004)} \\

31
& \mbox{\uposfeat{upos\_PROPN} (.024)}
& \mbox{\uposfeat{upos\_NOUN} (.015)}
& \mbox{\uposfeat{upos\_PRON} (.014)}
& \mbox{\uposfeat{upos\_ADV} (.011)}
& \mbox{\synfeat{dep\_nsubj} (.005)} \\

\rowcolor{black!5}[0pt][0pt]
\multicolumn{6}{@{}l@{}}{\textbf{Qwen2.5-7B}} \\
2
& \mbox{\rawfeat{n\_tokens} (.054)}
& \mbox{\uposfeat{upos\_PUNCT} (.012)}
& \mbox{\rawfeat{n\_sentences} (.011)}
& \mbox{\verbfeat{verbal\_root} (.010)}
& \mbox{\rawfeat{char\_per\_tok} (.005)} \\

14
& \mbox{\rawfeat{char\_per\_tok} (.018)}
& \mbox{\uposfeat{upos\_PRON} (.015)}
& \mbox{\synfeat{dep\_nsubj} (.006)}
& \mbox{\uposfeat{upos\_NOUN} (.006)}
& \mbox{\synfeat{dep\_obj} (.002)} \\

27
& \mbox{\verbfeat{verbal\_root} (.066)}
& \mbox{\uposfeat{upos\_PROPN} (.035)}
& \mbox{\uposfeat{upos\_NOUN} (.028)}
& \mbox{\rawfeat{char\_per\_tok} (.014)}
& \mbox{\synfeat{dep\_nsubj} (.004)} \\

\rowcolor{black!5}[0pt][0pt]
\multicolumn{6}{@{}l@{}}{\textbf{Apertus-8B}} \\
2
& \mbox{\rawfeat{n\_tokens} (.043)}
& \mbox{\rawfeat{n\_sentences} (.029)}
& \mbox{\verbfeat{verbal\_root} (.009)}
& \mbox{\rawfeat{char\_per\_tok} (.008)}
& \mbox{\uposfeat{upos\_NOUN} (.007)} \\

16
& \mbox{\uposfeat{upos\_NOUN} (.018)}
& \mbox{\rawfeat{char\_per\_tok} (.013)}
& \mbox{\synfeat{dep\_nsubj} (.011)}
& \mbox{\uposfeat{upos\_PRON} (.009)}
& \mbox{\rawfeat{n\_tokens} (.006)} \\

31
& \mbox{\uposfeat{upos\_PROPN} (.051)}
& \mbox{\uposfeat{upos\_NOUN} (.017)}
& \mbox{\synfeat{dep\_nsubj} (.016)}
& \mbox{\rawfeat{n\_tokens} (.013)}
& \mbox{\verbfeat{verbal\_root} (.006)} \\

\bottomrule
\end{tabularx}

\caption{Most predictive features for random forest classification of the lowest and highest ESS-estimated pointwise ID quintiles. Parentheses report the mean decrease in held-out ROC-AUC after permutation, averaged over eight shuffles and five cross-validation folds. Background colours indicate the four linguistic feature families: \familykey{rawfamily}{Raw text}, \familykey{uposfamily}{Morphosyntax}, \familykey{syntaxfamily}{Dependency and tree structure}, and \familykey{verbfamily}{Verbal and clausal structure}.}
\label{tab:ess-classification-features}
\end{table*}

\begin{table*}[t]
\centering
\small
\setlength{\tabcolsep}{4.5pt}
\renewcommand{\arraystretch}{1.10}

\begin{tabularx}{\textwidth}{
    @{}
    p{0.17\textwidth}
    p{0.14\textwidth}
    C{0.08\textwidth}
    C{0.075\textwidth}
    Y
    @{}
}
\toprule
\rowcolor{headerrow}
\textbf{Profiling-UD feature}
& \textbf{Association}
& \textbf{Feature value}
& \textbf{ESS ID}
& \textbf{Representative excerpt} \\
\midrule

\uposfeature{upos\_PROPN}
& High $\rightarrow$ high ID
& 19.5\%
& 179.520
& ``Any piece about \featurehl{New York City} has a heavyweight
champion to contend with-\featurehl{E.B. White}'s
\textit{Here Is New York}-but \featurehl{André Aciman}'s
\textit{Shadow Cities} comes out swinging.'' \\

\addlinespace[3pt]

\rawfeature{char\_per\_tok}
& High $\rightarrow$ low ID
& 9.75
& 43.780
& ``Plasma levels of \featurehl{trimethoprim} and
\featurehl{sulfonamide} after \featurehl{administration} of
\featurehl{trimethoprim-sulfamethoxazole} and
\featurehl{trimethoprim-sulfamoxole}.'' \\

\addlinespace[3pt]

\uposfeature{upos\_NOUN}
& High $\rightarrow$ low ID
& 48\%
& 50.740
& ``\featurehl{Determination} of \featurehl{tricarboxylic acid cycle
acids} and other related \featurehl{substances} in cultured mammalian
\featurehl{cells} by gradient \featurehl{ion-exchange chromatography}
with suppressed \featurehl{conductivity detection}.'' \\

\addlinespace[3pt]

\depfeature{dep\_nsubj}
& High $\rightarrow$ high ID
& 14.3\%
& 209.190
& ``Mine, too, \featurehl{I}'ve felt that \featurehl{there} is more
to the Yithians than most other mythos races\ldots{}
\featurehl{they}'re not just purely malicious or uncaring as so many
seem to be.'' \\

\addlinespace[3pt]

\verbalfeature{verbal\_root}
& High $\rightarrow$ high ID
& 100\%
& 186.820
& ``When I \featurehl{heard} that Children Of Bodom
\featurehl{was putting out} a new record, I \featurehl{was} both
excited and hesitant at the same time.'' \\

\addlinespace[3pt]

\rawfeature{n\_tokens}
& High $\rightarrow$ high ID
& 98
& 195.390
& ``We hear it from you all the time and we feel it too: community is
what we do best. We bring together a diverse group of creative people
who are living lives they love and YOU are part of that. You connect,
share, motivate, and inspire each other. You are movers \& shakers.
You are the ones who burn the midnight oil and wake up at the crack of
dawn. You save your money to invest in your future. You load up trucks
and package delicate things.'' \\

\addlinespace[3pt]

\uposfeature{upos\_PRON}
& High $\rightarrow$ high ID
& 16.7\%
& 198.890
& ``In other news: \featurehl{I}'ve re-discovered
\featurehl{my} love for Bismuth.'' \\

\bottomrule
\end{tabularx}

\caption{Representative excerpts illustrating the recurring feature directions identified by ALE. ESS-estimated pointwise ID values are for Pythia-6.9B at layer 16, with \(k=1000\) and \(n_{\mathrm{ref}}=15{,}000\). Teal highlights words most directly related to the corresponding feature. For \texttt{char\_per\_tok}, highlights identify representative long tokens; for \texttt{n\_tokens}, the complete excerpt illustrates input length. Examples are illustrative and do not isolate a feature from correlated linguistic properties. Background colours on feature labels indicate the four linguistic feature families: \familykey{rawfamily}{Raw text}, \familykey{uposfamily}{Morphosyntax}, \familykey{syntaxfamily}{Dependency and tree structure}, and \familykey{verbfamily}{Verbal and clausal structure}.}
\label{tab:mann-whitney}
\end{table*}

\section{Cross-fitted ALE estimation}
\label{app:ale}

Accumulated local effects (ALE) describe how a fitted regression model
responds to a feature over its observed range
\citep{apley2020visualizing}. We partition each feature into intervals
and, for every held-out observation in an interval, predict ID after
setting the feature to its lower and upper boundaries while retaining
the observed values of all other predictors. The prediction differences
are averaged within intervals and accumulated to form the ALE curve.
Because each interval uses only observations occurring within it, ALE
limits evaluation of poorly represented feature combinations.

Curves are estimated using the same five-fold cross-fitting protocol as
the predictive analysis. In each iteration, the random forest is fitted
on four folds and ALE is evaluated on the remaining fold. The held-out
effects are then combined. At \(k=1000\), we summarize ALE for the five
features with the largest reliably positive permutation importances in
each model, layer, and ID-estimator setting. An importance is considered
reliable when its mean is positive and its fold-level value is positive
in at least four folds. Model, imputation, and permutation settings are
reported in Appendix~\ref{app:hyp}.
For each curve, we calculate
\[
D_{10,90}
=
\operatorname{ALE}(q_{0.90})
-
\operatorname{ALE}(q_{0.10}),
\]
where \(q_{0.10}\) and \(q_{0.90}\) denote the feature's 10th and 90th
percentiles. The contrast is expressed in target standard-deviation
units: positive values indicate higher predicted ID toward the upper
end of the feature range, and negative values indicate lower predicted
ID. A direction is considered stable when its bootstrap interval
excludes zero and its sign agrees in at least four folds. Stable effects
are summarized in Table~\ref{tab:feature_direction}.

\section{Lexical baselines}\label{app:lexical}

We tested whether the results could be reproduced using word features
instead of the structured linguistic profile. Because the Profiling-UD
representation comprises 43 linguistic descriptors and three frequency
features, we constructed two matched lexical baselines, each containing
46 word-unigram TF-IDF features \citep{salton1988term,manning2008introduction}. The first used the most
frequent eligible words appearing in at least two and no more than 90\% of
the excerpts used in the main experiments. The second used frequent words appearing in 1 to
5\% of the same data. We chose the 1 to 5\% document frequency range as a pragmatic
moderate frequency band. The lower bound excludes highly sparse terms,
while the upper bound excludes common vocabulary already represented by the
frequent word baseline. The thresholds were fixed before evaluation and
were not tuned using the ID scores. Neither vocabulary was selected using the ID
scores. 
The text passages were lowercased and Unicode accents were removed. Vocabulary
selection, frequency filtering, TF-IDF weights, and normalization were fitted
separately within each cross-validation training fold. The held-out
documents were then transformed using only the vocabulary and weights
learned from that fold. For both regression and classification, we used random forests with 300 trees and a minimum leaf size of two, matching the settings of the Profiling-UD models.

The structured linguistic profile outperformed both lexical baselines in 26 of the 27 regression settings and in all 27 classification settings. The regression and classification comparisons are reported in Tables~\ref{tab:lexical_regression} and~\ref{tab:lexical_classification}, respectively. Overall, these results suggest that the 46-word lexical representations capture less predictive information than the broader linguistic profile. However, this is a deliberately small lexical baseline, so the result does not show that lexical information is unimportant or that the Profiling-UD signal is exclusively driven by abstract linguistic features.

\begin{table*}[!tp]
\begin{minipage}[t]{0.49\textwidth}
\captionsetup{hypcap=false}
\centering
\footnotesize
\setlength{\tabcolsep}{1.5pt}
\renewcommand{\arraystretch}{1.10}
\renewcommand{\tabularxcolumn}[1]{m{#1}}

\begin{tabularx}{\columnwidth}{l c *{4}{>{\centering\arraybackslash}X}}
\toprule
\shortstack{\textbf{ID}\\\textbf{target}} & \textbf{Layer} & \textbf{Linguistic \newline features}
& \textbf{Most\newline frequent} & \textbf{1-5\%\newline freq.}
& \textbf{Log\newline PPL} \\
\midrule
\rowcolor{black!7}
\multicolumn{6}{l}{\textbf{Pythia-6.9B}} \\
\multirow{3}{*}{ESS} & 2 & \textbf{.892} & .822 & .747 & .554 \\
 & 16 & \textbf{.958} & .911 & .855 & .754 \\
 & 31 & \textbf{.848} & .768 & .730 & .690 \\
\addlinespace[2pt]
\multirow{3}{*}{FCI} & 2 & \textbf{.881} & .824 & .754 & .531 \\
 & 16 & \textbf{.950} & .880 & .801 & .676 \\
 & 31 & \textbf{.800} & .686 & .680 & .545 \\
\addlinespace[2pt]
\multirow{3}{*}{MLE} & 2 & \textbf{.848} & .749 & .745 & .543 \\
 & 16 & \textbf{.900} & .794 & .723 & .642 \\
 & 31 & \textbf{.838} & .788 & .718 & .586 \\
\addlinespace[3pt]
\rowcolor{black!7}
\multicolumn{6}{l}{\textbf{Qwen2.5-7B}} \\
\multirow{3}{*}{ESS} & 2 & \textbf{.844} & .754 & .609 & .562 \\
 & 14 & \textbf{.976} & .926 & .865 & .848 \\
 & 27 & \textbf{.931} & .831 & .764 & .669 \\
\addlinespace[2pt]
\multirow{3}{*}{FCI} & 2 & \textbf{.824} & .737 & .655 & .526 \\
 & 14 & \textbf{.966} & .909 & .867 & .804 \\
 & 27 & \textbf{.925} & .820 & .705 & .558 \\
\addlinespace[2pt]
\multirow{3}{*}{MLE} & 2 & \textbf{.762} & .656 & .622 & .564 \\
 & 14 & \textbf{.877} & .798 & .741 & .724 \\
 & 27 & \textbf{.860} & .782 & .743 & .590 \\
\addlinespace[3pt]
\rowcolor{black!7}
\multicolumn{6}{l}{\textbf{Apertus-8B}} \\
\multirow{3}{*}{ESS} & 2 & \textbf{.864} & .777 & .650 & .576 \\
 & 16 & \textbf{.951} & .895 & .831 & .661 \\
 & 31 & \textbf{.869} & .824 & .743 & .547 \\
\addlinespace[2pt]
\multirow{3}{*}{FCI} & 2 & \textbf{.830} & .755 & .638 & .565 \\
 & 16 & \textbf{.941} & .883 & .843 & .652 \\
 & 31 & \textbf{.859} & .794 & .710 & .553 \\
\addlinespace[2pt]
\multirow{3}{*}{MLE} & 2 & \textbf{.836} & .679 & .630 & .545 \\
 & 16 & \textbf{.897} & .822 & .749 & .614 \\
 & 31 & \textbf{.876} & .783 & .739 & .545 \\
\bottomrule
\end{tabularx}

\caption{Classification performance of Profiling-UD, matched lexical baselines, and log perplexity on \texttt{main\_set}. Values are mean five-fold cross-validated ROC-AUC scores for random forest classifiers distinguishing excerpts in the lowest and highest pointwise ID quintiles, using identical excerpts and fold assignments within each setting. Boldface marks the highest displayed score in each row, including ties.}
\label{tab:lexical_classification}
\end{minipage}%
\hfill
\begin{minipage}[t]{0.49\textwidth}
\captionsetup{hypcap=false}
\centering
\footnotesize
\setlength{\tabcolsep}{1.5pt}
\renewcommand{\arraystretch}{1.10}
\renewcommand{\tabularxcolumn}[1]{m{#1}}

\begin{tabularx}{\columnwidth}{l c *{4}{>{\centering\arraybackslash}X}}
\toprule
\shortstack{\textbf{ID}\\\textbf{target}} & \textbf{Layer} & \textbf{Linguistic\newline features}
& \textbf{Most\newline frequent} & \textbf{1--5\%\newline freq.} & \textbf{Log\newline PPL} \\
\midrule

\rowcolor{black!7}
\multicolumn{6}{l}{\textbf{Pythia-6.9B}} \\

\multirow{3}{*}{ESS}
& 2 & \textbf{.256} & .154 & .068 & -.292 \\
& 16 & \textbf{.505} & .274 & .221 & -.077 \\
& 31 & \textbf{.263} & .135 & .104 & -.133 \\

\addlinespace[2pt]

\multirow{3}{*}{FCI}
& 2 & \textbf{.259} & .139 & .072 & -.235 \\
& 16 & \textbf{.424} & .192 & .155 & -.130 \\
& 31 & \textbf{.208} & .077 & .050 & -.217 \\

\addlinespace[2pt]

\multirow{3}{*}{MLE}
& 2 & \textbf{.127} & .007 & -.004 & -.183 \\
& 16 & \textbf{.259} & .072 & .073 & -.148 \\
& 31 & \textbf{.213} & .103 & .040 & -.137 \\

\addlinespace[3pt]
\rowcolor{black!7}
\multicolumn{6}{l}{\textbf{Qwen2.5-7B}} \\

\multirow{3}{*}{ESS}
& 2 & \textbf{.246} & .098 & -.017 & -.266 \\
& 14 & \textbf{.547} & .323 & .269 & .006 \\
& 27 & \textbf{.432} & .239 & .180 & -.105 \\

\addlinespace[2pt]

\multirow{3}{*}{FCI}
& 2 & \textbf{.204} & .070 & -.032 & -.254 \\
& 14 & \textbf{.482} & .294 & .222 & -.059 \\
& 27 & \textbf{.359} & .162 & .128 & -.197 \\

\addlinespace[2pt]

\multirow{3}{*}{MLE}
& 2 & \textbf{.049} & -.044 & -.074 & -.216 \\
& 14 & \textbf{.239} & .083 & .055 & -.116 \\
& 27 & \textbf{.159} & .145 & .025 & -.210 \\

\addlinespace[3pt]
\rowcolor{black!7}
\multicolumn{6}{l}{\textbf{Apertus-8B}} \\

\multirow{3}{*}{ESS}
& 2 & \textbf{.270} & .130 & -.031 & -.063 \\
& 16 & \textbf{.439} & .258 & .199 & -.075 \\
& 31 & \textbf{.311} & .193 & .121 & -.114 \\

\addlinespace[2pt]

\multirow{3}{*}{FCI}
& 2 & \textbf{.253} & .131 & -.039 & -.070 \\
& 16 & \textbf{.421} & .239 & .183 & -.073 \\
& 31 & \textbf{.286} & .173 & .103 & -.127 \\

\addlinespace[2pt]

\multirow{3}{*}{MLE}
& 2 & \textbf{.079} & -.028 & -.049 & -.114 \\
& 16 & \textbf{.268} & .118 & .088 & -.075 \\
& 31 & .101 & \textbf{.110} & .016 & -.124 \\

\addlinespace[2pt]
\bottomrule
\end{tabularx}

\caption{Regression performance of Profiling-UD, matched lexical baselines, and log perplexity on \texttt{main\_set}. Values are mean five-fold cross-validated \(R^2\) scores for random forest regression on each pointwise ID target, using identical excerpts and folds within each setting. Boldface marks the highest displayed score in each row, including ties.}
\label{tab:lexical_regression}
\end{minipage}%
\end{table*}

\section{ID profiles across checkpoints}
\label{app:pythia_pretraining}

\begin{figure*}[!tp]
    \centering
    \includegraphics[width=\textwidth]{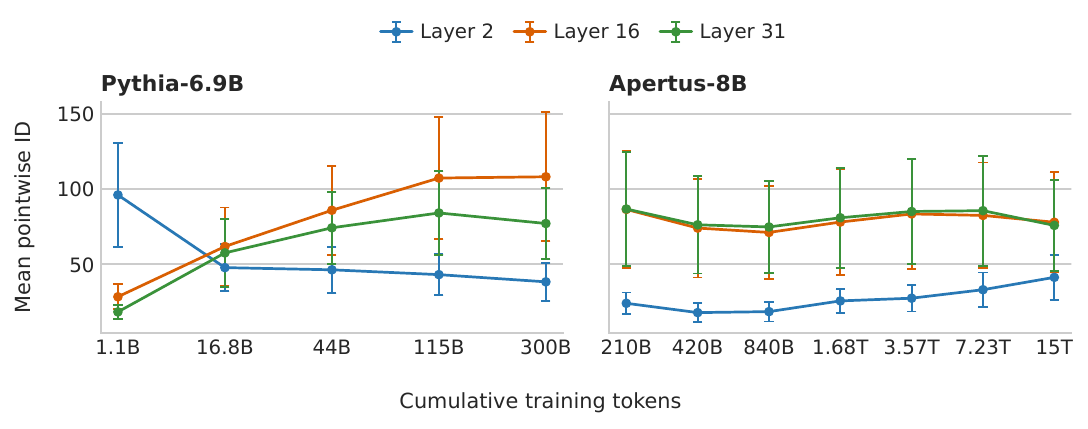}
\caption{Mean ESS-estimated pointwise ID (\(\pm1\sigma\), where \(\sigma\) is the standard deviation across excerpts) at early, intermediate, and late layers across Pythia and Apertus pretraining checkpoints. The x-axes report approximate cumulative training tokens (B: billion; T: trillion). Checkpoints are equally spaced within each model; the two training trajectories are not temporally aligned.}
\label{fig:ess-checkpoints}
\end{figure*}

To place the checkpoint-level prediction results in their broader geometric context, Figure~\ref{fig:ess-checkpoints} shows how mean ESS-estimated pointwise ID changes across training. The trajectories must be separately interpreted within each model, because the Pythia and Apertus checkpoints are not temporally aligned.

Pythia complements the Apertus analysis by providing checkpoints much closer to initialization. The largest change occurs between \texttt{step\_512} and \texttt{step\_8000}. At the latter checkpoint,
random forest performance reaches \(R^2=0.479\) at layer 16 and
\(0.396\) at layer 31, and the importance rankings begin to include more structural linguistic properties (Table~\ref{tab:pythia-feature}). Sentence count is temporarily prominent at this stage, whereas punctuation remains concentrated in the early layer. At subsequent checkpoints, intermediate and late layers increasingly emphasize nominal composition, pronouns, nominal subjects, and verbal roots. The intermediate-layer advantage also remains clear. These results suggest that the depth-specific relationship between linguistic properties and pointwise ID emerges early in Pythia and is subsequently refined rather than reorganized completely.

\begin{table*}[!tp]
\centering
{\scriptsize
\setlength{\tabcolsep}{3pt}
\setlength{\fboxsep}{0.8pt}
\renewcommand{\arraystretch}{1.03}

\begin{tabularx}{\textwidth}{
    @{}
    l
    c
    r
    >{\raggedright\arraybackslash}X
    @{}
}
\toprule
\textbf{Checkpoint}
& \textbf{Layer}
& \textbf{ESS \(R^2\)}
& \textbf{Five largest permutation importances
  \((\Delta R^2)\)} \\
\midrule

\multirow{3}{*}{\texttt{step\_512}}
& 2
& .146
& \uposfeat{upos\_PUNCT} (.081);
  \uposfeat{upos\_PRON} (.027);
  \frequencyfeat{median\_frequency\_class} (.021);
  \rawfeat{char\_per\_tok} (.021);
  \verbalfeat{verbal\_root} (.014) \\

& 16
& .197
& \rawfeat{char\_per\_tok} (.067);
  \uposfeat{upos\_PUNCT} (.035);
  \uposfeat{upos\_DET} (.024);
  \uposfeat{upos\_PRON} (.018);
  \verbalfeat{aux\_tense\_dist\_Pres} (.015) \\

& 31
& .183
& \rawfeat{char\_per\_tok} (.037);
  \uposfeat{upos\_PUNCT} (.028);
  \uposfeat{upos\_DET} (.022);
  \uposfeat{upos\_NUM} (.021);
  \uposfeat{upos\_ADJ} (.019) \\

\addlinespace[2pt]

\multirow{3}{*}{\texttt{step\_8000}}
& 2
& .167
& \uposfeat{upos\_PUNCT} (.110);
  \uposfeat{upos\_PROPN} (.059);
  \uposfeat{upos\_PART} (.013);
  \rawfeat{tokens\_per\_sent} (.013);
  \uposfeat{upos\_DET} (.011) \\

& 16
& .479
& \rawfeat{n\_sentences} (.584);
  \uposfeat{upos\_PROPN} (.044);
  \rawfeat{char\_per\_tok} (.040);
  \uposfeat{upos\_NOUN} (.039);
  \syntaxfeat{dep\_nsubj} (.039) \\

& 31
& \textbf{.396}
& \rawfeat{n\_sentences} (.477);
  \verbalfeat{verbal\_root} (.058);
  \uposfeat{upos\_PROPN} (.045);
  \syntaxfeat{dep\_nsubj} (.033);
  \uposfeat{upos\_NOUN} (.024) \\

\addlinespace[2pt]

\multirow{3}{*}{\texttt{step\_21000}}
& 2
& .199
& \uposfeat{upos\_PUNCT} (.071);
  \rawfeat{n\_tokens} (.071);
  \uposfeat{upos\_PROPN} (.043);
  \uposfeat{upos\_NOUN} (.021);
  \uposfeat{upos\_NUM} (.012) \\

& 16
& .522
& \rawfeat{char\_per\_tok} (.121);
  \uposfeat{upos\_PRON} (.099);
  \uposfeat{upos\_PROPN} (.084);
  \uposfeat{upos\_NOUN} (.072);
  \verbalfeat{verbal\_root} (.038) \\

& 31
& .291
& \verbalfeat{verbal\_root} (.062);
  \uposfeat{upos\_PROPN} (.059);
  \uposfeat{upos\_PRON} (.043);
  \uposfeat{upos\_NOUN} (.038);
  \syntaxfeat{dep\_nsubj} (.035) \\

\addlinespace[2pt]

\multirow{3}{*}{\texttt{step\_55000}}
& 2
& .184
& \rawfeat{n\_tokens} (.060);
  \uposfeat{upos\_PUNCT} (.052);
  \uposfeat{upos\_NOUN} (.021);
  \uposfeat{upos\_PROPN} (.018);
  \rawfeat{char\_per\_tok} (.013) \\

& 16
& \textbf{.523}
& \uposfeat{upos\_NOUN} (.176);
  \rawfeat{char\_per\_tok} (.083);
  \uposfeat{upos\_PRON} (.074);
  \uposfeat{upos\_PROPN} (.058);
  \verbalfeat{verbal\_root} (.054) \\

& 31
& .284
& \verbalfeat{verbal\_root} (.082);
  \uposfeat{upos\_NOUN} (.071);
  \uposfeat{upos\_PROPN} (.052);
  \uposfeat{upos\_PRON} (.031);
  \uposfeat{upos\_ADV} (.019) \\

\addlinespace[2pt]

\multirow{3}{*}{\texttt{final/main}}
& 2
& \textbf{.256}
& \rawfeat{n\_tokens} (.159);
  \uposfeat{upos\_PROPN} (.041);
  \uposfeat{upos\_PUNCT} (.033);
  \syntaxfeat{dep\_nsubj} (.020);
  \rawfeat{char\_per\_tok} (.018) \\

& 16
& .505
& \uposfeat{upos\_NOUN} (.149);
  \rawfeat{char\_per\_tok} (.091);
  \uposfeat{upos\_PRON} (.074);
  \uposfeat{upos\_PROPN} (.067);
  \verbalfeat{verbal\_root} (.045) \\

& 31
& .263
& \uposfeat{upos\_NOUN} (.075);
  \uposfeat{upos\_PROPN} (.061);
  \verbalfeat{verbal\_root} (.040);
  \uposfeat{upos\_PRON} (.032);
  \syntaxfeat{dep\_nsubj} (.020) \\

\bottomrule
\end{tabularx}
}
\caption{ESS-estimated pointwise ID regression across five Pythia checkpoints. Scores are mean five-fold cross-validated \(R^2\) values for random forest regression. For each layer, boldface marks the highest score across checkpoints. Parentheses report the mean decrease in held-out \(R^2\) after permutation, averaged over eight shuffles and five cross-validation folds. Background colours indicate the linguistic feature families and frequency controls: \familykey{rawfamily}{Raw text}, \familykey{uposfamily}{Morphosyntax}, \familykey{syntaxfamily}{Dependency and tree structure}, \familykey{verbfamily}{Verbal and clausal structure}, and \familykey{frequencyfamily}{Lexical frequency}.}
\label{tab:pythia-feature}
\end{table*}

\onecolumn
\section{Linguistic features}\label{app:features}
We retain 43 Profiling-UD features (Table~\ref{tab:retained_profiling_ud_features}) and augment them with three tokenizer-specific lexical-frequency statistics, yielding 46 predictors in the full linguistic-profiling model. Profiling-UD token counts refer to UD word tokens, whereas the frequency-class features are computed over each model's native subword tokenization. Morphological, part-of-speech, and dependency labels follow Universal Dependencies \citep{nivre2020universal}.

\begingroup
\centering
\footnotesize
\setlength{\tabcolsep}{5pt}
\renewcommand{\arraystretch}{1.00}

\begin{tabularx}{\textwidth}{
    @{}
    >{\ttfamily\raggedright\arraybackslash}p{0.34\textwidth}
    >{\raggedright\arraybackslash}X
    @{}
}
\toprule
\multicolumn{1}{@{}l}{\textbf{Feature}}
&
\multicolumn{1}{l@{}}{\textbf{Definition}} \\
\midrule

\multicolumn{2}{@{}l}{%
  \textsc{Profiling-UD descriptors}
  \enspace\textit{(43 predictors)}
} \\[-0.25ex]

\featgroup{Surface and lexical composition}{4}

n\_sentences
& Number of sentences detected by the UD parser. \\

n\_tokens
& Number of UD word tokens, distinct from the number of
language-model subword tokens. \\

tokens\_per\_sent
& Mean number of UD word tokens per parser-detected sentence. \\

char\_per\_tok
& Mean number of characters per non-punctuation UD token. \\

\featgroup{Dependency-tree geometry}{3}

avg\_links\_len
& Mean absolute head-dependent distance in UD tokens, excluding
punctuation dependencies. \\

max\_links\_len
& Maximum head-dependent distance, using the same dependency
filtering. \\

avg\_max\_depth
& Mean maximum root-to-leaf dependency-tree depth across
parser-detected sentences. \\

\featgroup{Clausal and verbal-predicate structure}{6}

avg\_prepositional\_chain\_len
& Mean depth of recursively embedded prepositional or adpositional
modifier chains. \\

avg\_subordinate\_chain\_len
& Mean depth of recursively embedded subordinate-clause chains. \\

avg\_token\_per\_clause
& Mean number of UD tokens per clause. \\

avg\_verb\_edges
& Mean number of eligible dependency links attached to each verbal
head, excluding punctuation and copular links. \\

subordinate\_proposition\_dist
& Percentage of clauses classified as subordinate rather than main
clauses. \\

verbal\_root
& Percentage of parser-detected sentences whose dependency root is
tagged as \texttt{VERB}. \\

\featgroup{Linear order}{3}

subj\_pre
& Percentage of subject dependents preceding their governing verbal
head. \\

obj\_post
& Percentage of core-object dependents following their governing
verbal head. \\

subordinate\_post
& Percentage of subordinate-clause heads following their governing
head. \\

\featgroup{Dependency-relation distributions}{5}

dep\_\{compound, nmod, nsubj, obj, obl\}
& Five predictors giving the percentages of UD dependency relations
labelled as compound, nominal modifier, nominal subject, core object,
and oblique nominal. \\

\featgroup{Morphosyntactic information}{14}

upos\_\{ADJ, ADP, ADV, AUX, CCONJ, DET,\newline
NOUN, NUM, PART, PRON, PROPN, PUNCT, VERB\}
& Thirteen predictors giving the percentage of UD tokens assigned
each listed universal part-of-speech tag. \\

lexical\_density
& Percentage of word tokens assigned a content-word UPOS tag:
\texttt{NOUN}, \texttt{VERB}, \texttt{ADJ}, or \texttt{ADV}. \\
\featgroup{Auxiliary morphology}{3}

aux\_mood\_dist\_Ind;\newline
aux\_num\_pers\_dist\_Sing+3;\newline
aux\_tense\_dist\_Pres
& Among tokens tagged as \texttt{AUX}, the percentages marked
\texttt{Mood=Ind}, jointly marked \texttt{Number=Sing} and
\texttt{Person=3}, and marked \texttt{Tense=Pres}, respectively. \\

\featgroup{Lexical-verb morphology}{5}

verbs\_form\_dist\_\{Inf, Part\}
& Two predictors giving the percentages of \texttt{VERB} tokens
marked as infinitives and participles. \\

verbs\_mood\_dist\_Ind
& Percentage of \texttt{VERB} tokens marked for indicative mood. \\

verbs\_tense\_dist\_\{Past, Pres\}
& Two predictors giving the percentages of \texttt{VERB} tokens
marked for past and present tense. \\

\midrule

\multicolumn{2}{@{}l}{%
  \textsc{Tokenizer-specific frequency controls}
  \enspace\textit{(3 predictors)}
} \\[-0.25ex]

\addlinespace[2pt]

median\_frequency\_class
& Median frequency class \(C_M(t)\) across the model tokens in the
input. \\

max\_frequency\_class
& Largest frequency class in the input, corresponding to its rarest
token under the model-specific frequency table. \\

min\_frequency\_class
& Smallest frequency class in the input, corresponding to its most
frequent token under the model-specific frequency table. \\

\bottomrule
\end{tabularx}

\captionsetup{hypcap=false}
\captionof{table}{Inventory of the 46 predictors used in the pointwise ID analyses. The first 43 are Profiling-UD descriptors grouped by linguistic level; the final three summarize tokenizer-specific frequency classes estimated from The Pile.}
\label{tab:retained_profiling_ud_features}
\par
\endgroup

\end{document}